\documentclass[journal,twoside,web]{ieeecolor}
\usepackage{generic}
\usepackage{cite}
\usepackage[caption=false,font=normalsize,labelfont=sf,textfont=sf]{subfig}
\usepackage{amsmath,amssymb,amsfonts}
\usepackage{graphicx}
\usepackage{algorithm}
\usepackage{algpseudocode}
\usepackage{array}
\usepackage{hyperref}
\usepackage{makecell}
\usepackage{booktabs}
\usepackage{multirow}
\usepackage[table]{xcolor}   
\usepackage{graphicx}        
\usepackage{textcomp}
\usepackage{tcolorbox}
\usepackage{tikz}
\usepackage[normalem]{ulem} 
\usepackage{xcolor,pifont}
\newcommand{\cmark}{\textcolor{green!60!black}{\ding{52}}}
\newcommand{\xmark}{\textcolor{red}{\ding{55}}}

\newcommand{\figref}[1]{Fig.~\ref{fig:#1}}

\newcommand{\tblref}[1]{Table~\ref{tab:#1}}

\def\BibTeX{{\rm B\kern-.05em{\sc i\kern-.025em b}\kern-.08em
    T\kern-.1667em\lower.7ex\hbox{E}\kern-.125emX}}
\newcolumntype{C}[1]{>{\centering\arraybackslash}p{#1}}
\begin{document}
\title{GlyRAG: Context-Aware Retrieval-Augmented Framework for Blood Glucose Forecasting}
\author{Shovito Barua Soumma$^{1,2}$,~\IEEEmembership{Student Member, IEEE}, Hassan Ghasemzadeh$^{1}$,
~\IEEEmembership{Senior Member, IEEE}
\thanks{
$^{1}$College of Health Solutions, Arizona State University, Phoenix, AZ 85004, USA. 
Emails: \textcolor{blue}{\{shovito, hassan.ghasemzadeh\}@asu.edu}.
}
\thanks{
$^{2}$School of Computing and Augmented Intelligence, Arizona State University, Tempe, AZ 85281, USA.
}
\thanks{This work was supported in part by the National Science Foundation (NSF) under grant 2402650. The content is the responsibility of the authors and does not necessarily represent the official views of the NSF.}
}

\maketitle

\begin{abstract}
\textcolor{black}
{Accurate blood glucose forecasting using continuous glucose monitoring (CGM) data can support early prediction of dysglycemic risk. However, current neural-network-based forecasting models treat CGM data as a purely numerical sequence without integrating contextual information that the CGM signal morphology contains. Recently, large language models (LLMs) have shown promise for time-series forecasting tasks, yet their role as agentic context extractors in diabetes care remains largely unexplored. In this study, we bridge glucose forecasting and LLM-based contextualization, and develop GlyRAG, a context-aware retrieval-augmented forecasting framework that uses an LLM as a contextualization agent to summarize glucose morphology directly from a timed CGM window. The generated CGM-only narrative is embedded and fused with patch-based glucose representations, and a retrieval module incorporates similar historical training episodes through cross-attention. We evaluate GlyRAG on OhioT1DM and AZT1D datasets for 5-, 30-, and 60-minute forecasting horizons. Against strong CGM-only baselines, GPT-4 GlyRAG significantly improves long-horizon RMSE (Root Mean Square Error) over PatchTST on both datasets. For example, RMSE decreases from 13.8 to 10.6 at 30 minutes and from 23.1 to 20.2 at 60 minutes on OhioT1DM. LLaMA-3.1 shows smaller but significant long-horizon gains, suggesting that the contextualization pipeline is not limited to GPT-4. Clinical error-grid analyses further show that approximately 85\% of predictions fall in clinically acceptable Clarke Error Grid Zones A--B. These results suggest that CGM-derived linguistic context and case-based retrieval can improve long-horizon glucose forecasting without requiring additional sensing modalities.}

\end{abstract}


\section{Introduction}

\IEEEPARstart{D}{iabetes} mellitus is a chronic metabolic disorder affecting hundreds of millions of adults worldwide and its prevalence continues to rise, driven largely by aging populations, sedentary lifestyles, and unhealthy diets. Type 2 diabetes (T2D) accounts for the majority of cases and is closely linked to other chronic conditions such as obesity and cardiovascular disease~\cite{WHOdiabetes2020}. In contrast, type 1 diabetes (T1D) is an autoimmune disease in which destruction of pancreatic $\beta$‑cells leads to lifelong dependence on exogenous insulin~\cite{diabetes2005intensive, karvonen2000incidence}. Globally, over 537 million adults live with diabetes, while an estimated 240 million remain undiagnosed for 4--7 years before detection~\cite{idf2021}. Poor glycemic control—with glucose frequently outside the target range of roughly 70–180 mg/dL—substantially increases the risk of micro and macrovascular complications, including cardiovascular disease, kidney failure, retinopathy, and neuropathy, underscoring the need for tight day‑to‑day glucose management in patients with diabetes~\cite{fullerton2014intensive,mccoy2017trajectories}.

Accurate forecasting of future blood glucose levels is critical in maintaining safe and stable glycemic control in diabetes management because glucose forecasting allows for early interventions to prevent abnormal glucose events such as hyperglycemia and hypoglycemia. Short-term predictions (e.g., 15--30 minutes prediction horizon) enable timely interventions such as insulin dose adjustments, carbohydrate supplementation, and activity modifications to mitigate impending dysglycemic episodes. However, extending prediction horizons (e.g., $\geq$60 minutes) provides additional lead time to adjust therapy or behavior, with the potential to reduce acute events and the cognitive burden of constant manual decision making~\cite{marigliano2024glucose}. Errors at extreme glucose ranges carry the greatest clinical risk, whereas modest deviations within the euglycemic band are less critical~\cite{woldaregay2019data}. 


Continuous glucose monitoring (CGM) has emerged as a key enabling technology for such forecasting. Modern CGM systems provide dense, near-real-time glucose measurements and are increasingly used not only by individuals with T1D but also by those with T2D and even by people with prediabetes or at-risk individuals seeking to understand and improve their metabolic health~\cite{kwon2025advances,cappon2019continuous}. Recent regulatory changes and over‑the‑counter availability have made CGM more accessible to the general public, creating an opportunity to leverage these data streams for personalized feedback and lifestyle guidance. In this setting, robust blood glucose forecasting models can serve as decision‑support tools to maintain time‑in‑range, prevent extreme excursions, and promote healthier day‑to‑day behaviors across the spectrum from T1D to T2D and prediabetes~\cite{arefeen2025glytwin}.

\textbf{Prior approaches:} Early work on glucose prediction compared classical time‑series models (e.g., autoregressive and state‑space methods) with data‑driven machine learning, showing that learned models generally outperform linear baselines, especially at longer horizons~\cite{xie2020}. Recurrent neural networks, most notably LSTMs (Long Short-Term Memory) and their variants, then became the dominant paradigm for sequential CGM forecasting, often augmented with ancillary inputs such as insulin and carbohydrate records~\cite{mujahid2021}. More recently, transformer-based architectures and multitask learning frameworks have demonstrated strong performance by leveraging self-attention mechanisms and multimodal covariates to extend predictive horizons and improve temporal representation~\cite{hwang2025generalized,machiraju2025timeaware}.

\textbf{Key limitations:} Despite recent progress, most existing models treat CGM primarily as a numerical sequence and either ignore contextual information or encode it in a narrow, hand‑engineered manner. Behavioral and clinical drivers—such as meals, activity, stress, heart rate, illness, and therapy changes—are only partially observed, and collecting rich multimodal data from multiple sensors at scale is challenging, making context‑heavy models difficult to train and deploy in routine care~\cite{machiraju2025timeaware, li2019glunet, shuvo2023deep}. Recent work in other time‑series domains shows that foundation models and multimodal agents can enhance accuracy and interpretability via contextual summaries and semantic reasoning, but their application to CGM forecasting remains limited by difficulties in extracting clinically meaningful context and designing efficient retrieval mechanisms~\cite{das2024decoder, jin2024timellm}. These gaps motivate frameworks that derive context understanding and semantic reasoning directly from CGM data, enabling models that rely solely on glucose traces to predict future values while capturing the underlying physiological state transitions that drive them. 
\textcolor{black}{These gaps motivate the development of new forecasting frameworks that derive contextual representations directly from CGM traces and use them to support longer-horizon forecasting without requiring additional sensing modalities.}

\begin{figure}[]
\centering
\vspace{-3mm}
\includegraphics[width=1\linewidth,trim={140 220 250 200},clip]{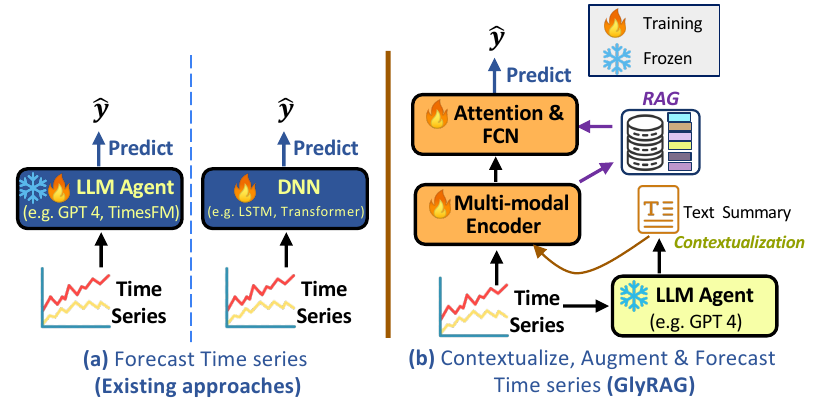}
\caption{Approaches of time‑series forecasting using LLM: (a) Existing methods either use LLM directly on CGM time series or depend on other sensors. (b) GlyRAG first uses an LLM agent to contextualize CGM windows, then fuses text and signal embeddings with retrieval‑augmented attention for forecasting.}
\vspace{-3mm}
\label{fig:motivation}
\end{figure}

\textbf{Novel contributions:} To address the above-mentioned gaps, we propose GlyRAG, a context‑aware, retrieval‑augmented agentic forecasting framework designed specifically for accurate blood glucose forecasting. In contrast to prior work that either ignores context or depends on additional sensors, GlyRAG derives semantic context directly from glucose traces and uses it to guide long‑horizon forecasting as shown in ~\figref{motivation}. Our main contributions are:
\begin{itemize}
    \item \textbf{\textcolor{black}{CGM-only} Context Extraction:} We use LLMs as contextualization agents to generate morphology‑aware, clinically meaningful textual summaries from blood-glucose windows and project them back into the forecasting pipeline, enabling context‑aware prediction without requiring extra modalities (e.g., wearables or additional clinical sensors).
    \item  \textbf{Retrieval‑augmented (RAG) forecasting: }We introduce a retrieval module that searches a library of historical blood-glucose embeddings for similar patterns and fuses them via cross‑attention, operationalizing case‑based reasoning.
    \item \textbf{Clinical validation:} We further perform retrospective clinical evaluation (e.g., clarke error grid, time-in-range etc.) 
    on two large-scale datasets, showing that GlyRAG delivers clinically reliable long‑horizon forecasts for decision support. 
\end{itemize}


\textcolor{black}{Across two T1D cohorts, our experiments show that GlyRAG improves long-horizon forecasting over strong CGM-only baselines, while ablation analyses clarify the separate contributions of LLM-derived context and retrieval.}

\begin{figure*}[t]
\vspace{-5mm}
\centering
\includegraphics[width=0.95\linewidth,trim={8 8 8 6},clip]{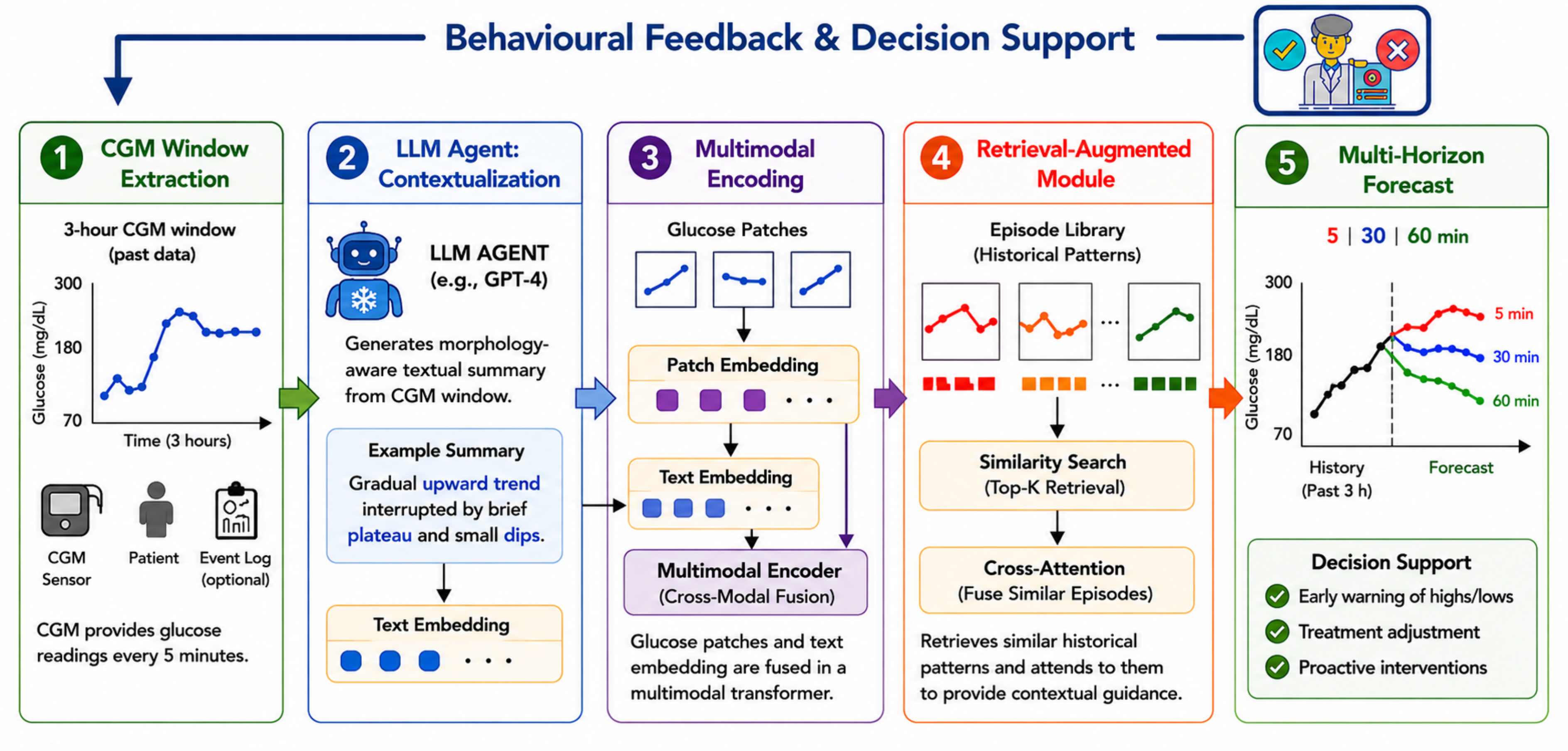}
\vspace{-3mm}
\caption{\color{black}\textbf{Overview of the proposed GlyRAG pipeline}: Three‑hour CGM windows are extracted and summarized by an LLM agent into short morphology‑aware text, which is embedded and fused with glucose patches in a multimodal encoder. A retrieval‑augmented module then attends to similar historical patterns to generate multi‑horizon forecasts that can be used for behavioral feedback and decision support.}
\label{fig:method}
\vspace{-5mm}
\end{figure*}

\section{Related Work}
\label{sec:relatedworks}
\subsection{Deep Learning \& Multimodal Approaches}
Blood glucose forecasting has progressed from classical autoregressive models to deep neural architectures.  Early LSTM-based approaches dominated sequential CGM modeling by capturing temporal dependencies~\cite{shuvo2023deep}, while hybrid CNN-LSTM methods combined local pattern extraction with recurrent processing~\cite{freiburghaus2020deep, jaloli2023long}.
Several works have explicitly injected clinical knowledge into glucose prediction pipelines. De Bois et al. incorporate regulatory criteria into training via a clinically weighted loss (gcMSE) and a progressive optimization scheme, trading raw accuracy for improved clinical acceptability of predictions~\cite{de2021integration}. Prendin et al. use explainable AI (SHAP) to reveal when similarly accurate LSTM models differ in physiological plausibility, arguing that interpretability is essential for safe decision support~\cite{prendin2023importance}, while Marigliano et al. demonstrate that CGM systems with predictive hypoglycemia alarms can materially improve time-below-range, underscoring the need for forecasting models that are both accurate and clinically aligned~\cite{marigliano2024glucose}.

Recognizing that glucose dynamics depend on multiple factors, several studies have incorporated additional modalities. GluNet integrates CGM with insulin delivery and carbohydrate intake, while deep multitask LSTM frameworks jointly model glucose trajectories with meals and exercise streams to improve forecasting and hypoglycemia detection~\cite{li2019glunet, daniels2021multitask}. More recent multimodal systems—such as Hwang et al.’s DA‑CMTL framework~\cite{hwang2025generalized}, time‑aware cross‑attention model~\cite{machiraju2025timeaware}, and GlucoNet~\cite{farahmand2025hybridattentionmodelusing}—fuse CGM with physiological and behavioral signals (e.g., heart rate, accelerometry, electrodermal activity) and hybrid transformer architectures, achieving strong accuracy but at the cost of dense multi-sensor logging, increased user burden, and greater system complexity, which limit deployment in routine care.

\subsection{LLM for Time-Series Forecasting (TSF)}
The success of large language models in natural language processing has inspired their application to TSF. 
Models such as TimeGPT~\cite{garza2023timegpt1}, Time-LLM~\cite{jin2024timellm}, and TimesFM~\cite{das2024decoder} show that pretrained transformers can be adapted to diverse forecasting tasks via reprogramming or few-shot prompting, and achieve competitive performance on standard benchmarks. However, these approaches typically operate on raw time-series values or learned numeric embeddings that differ substantially from the textual distributions on which LLMs are trained, limiting the extent to which they can exploit rich semantic priors.

Recent work has tried to bridge this gap by “textualizing’’ time series and adding simple metadata as prompts in a zero‑shot setting, but the resulting context is often shallow and hand‑crafted~\cite{xue2023promptcast, liu2023large}.

Building on this idea, TimeCAP introduces dual LLM agents and a multimodal encoder that contextualize, augment, and then predict discrete events from time series, showing sizable gains for classification tasks in weather, finance, and aggregated healthcare signals~\cite{lee2025timecap}. Yet, these frameworks still treat LLMs primarily as predictors and have not been adapted to patient‑level blood-glucose forecasting, where long, continuous trajectories must remain clinically plausible over 30–60 minutes and beyond. Moreover, these foundation models often struggle to generalize to specialized domains like healthcare. In our experiments, general-purpose models such as TimesFM also show limited generalizability to CGM data (~\tblref{clinical_metrics}), underperforming domain-specific baselines for long-horizon BGL prediction.

Unlike TimesFM’s task-agnostic forecasting and TimeCAP’s event-classification focus, GlyRAG uses the LLM as an \textit{agentic contextualization module} that guides multi‑horizon prediction from CGM alone—addressing the underexplored need for context‑aware BGL forecasting in healthcare. To the best of our knowledge, this is the first work to deploy an LLM‑based contextual and retrieval framework for long‑horizon CGM forecasting.

\section{Materials and Methods}
\subsection{Problem Formulation}
Let $\mathbf{x}$ denote blood glucose readings captured over a time window of length $L$ using a continuous glucose monitor (CGM):

\[
\mathbf{x} = (x_1, x_2, \ldots, x_{_L}), \quad x_t \in \mathbb{R}.
\]
where $x_t$ denotes the glucose value at time step $t$, and $L$ is the length of the input signal segment (i.e., time window). The forecasting task aims to predict future glucose values over a horizon $H$, producing estimates
\[
\mathbf{\hat{y}}=(\hat{y}_{_{L+1}},\hat{y}_{_{L+2}},\ldots,\hat{y}_{_{L+H}})
\]
to approximate the ground truth sequence $\mathbf{y}=(y_{_{L+1}},y_{_{L+2}},\ldots,y_{_{L+H}})$. We formulate this task as a supervised sequence-to-sequence regression problem, where the forecasting loss is measured by
the Huber Loss with $\delta = 1.0$:
\begin{equation}
    \mathcal{L}{_\text{forecast}} = \frac{1}{H} \sum_{h=1}^{H}\ell_{\delta}\big( \hat{y}_{_{L+h}} - y_{_{L+h}} \big),
\label{eq:huber_loss}
\end{equation}
where the Huber function is given by:
\begin{equation}
\ell_{\delta}(e) = 
\begin{cases}
\frac{1}{2}e^2 & \text{if } |e| \leq \delta \\
\delta \left( |e| - \frac{1}{2}\delta \right) & \text{if } |e| > \delta
\end{cases}
\label{eq:huber_function}
\end{equation}
and $e = \hat{y}_{_{L+h}} - y_{_{L+h}}$ represents the prediction error at the forecast step $h$.
We choose the Huber loss because it provides robustness to outliers commonly present in CGM data due to sensor noise and physiological anomalies~\cite{gokcesu2021generalizedhuberlossrobust}.



Unlike conventional time-series forecasting that relies solely on numerical sequences, 
blood glucose dynamics exhibit complex patterns influenced by physiological context—including glycemic variability, directional trends, and rate-of-change characteristics—that are not directly captured in raw CGM values. We formulate this as a multimodal learning problem where contextual understanding augments numerical pattern recognition and our framework incorporates contextual signals derived from large language models (LLMs) to augment the time series and improve predictive robustness.
\begin{figure}[]
\centering
\includegraphics[width=1\linewidth,trim={200 140 200 140},clip]{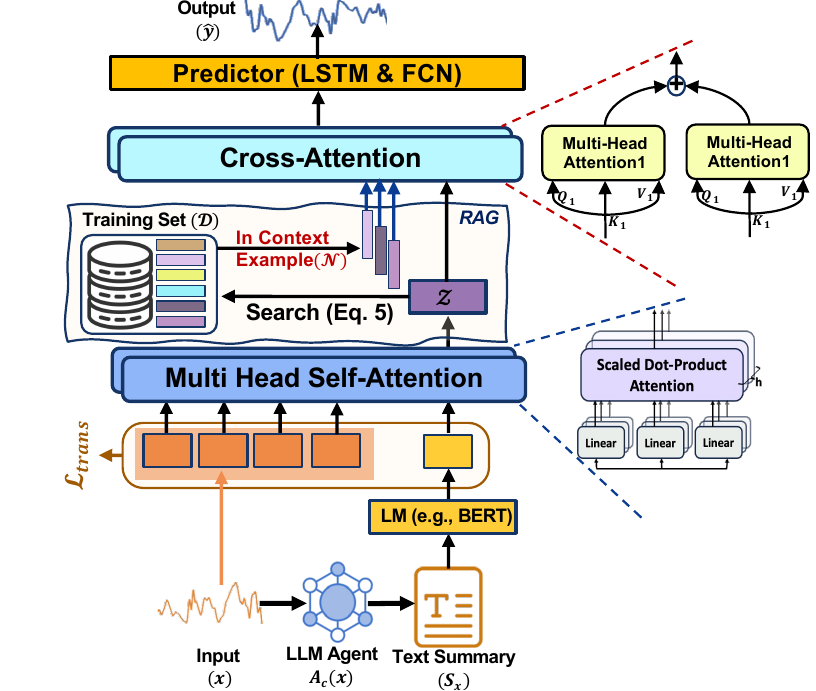}
\vspace{-5mm}
\caption{\textbf{Overall GlyRAG architecture:} (a) An LLM agent generates a morphology‑aware text summary from the input CGM window, which is encoded by a language model and fused with patch‑based glucose embeddings in a multi‑head self‑attention encoder to produce a joint context–CGM representation ($z$). (b) The fused query embedding searches a retrieval index for K similar historical episodes; cross‑attention branches combine the query with each neighbor, and the aggregated retrieval‑aware representation is fed to a predictor to forecast.}
\vspace{-3mm}
\label{fig:architecture}
\end{figure}
\subsection{Framework Overview}
As shown in \figref{architecture}, GlyRAG consists of three main components: (1) an LLM-based context generator that produces textual descriptions of glucose dynamics such as rising trends, sharp drops, or oscillations, (2) a multimodal transformer (MMT) encoder that fuses context and time-series representations through cross-attention with alignment constraints, and (3) a retrieval-augmented forecasting module that leverages similar historical patterns during inference. 
An overview of our framework—from CGM windowing and LLM‑based contextualization to retrieval‑augmented multimodal forecasting and behavioral feedback—is shown in \figref{method}, and the internal architecture of the multimodal encoder and retrieval adapter is detailed in \figref{architecture}. \textcolor{black}{
Here, the term contextualization agent refers only to the LLM module that converts an observed CGM window into a morphology-aware summary. GlyRAG does not perform autonomous treatment planning, insulin/carbohydrate recommendations, or closed-loop decision-making; its output remains a glucose forecast.
}


\subsection{LLM-Based Context Generation}
We employ a large language model $M_\theta$ (e.g., GPT-4) as a contextualization agent $A_C$ to generate textual summaries of glucose morphology. Given an input CGM sequence $\mathbf{x}$, the agent produces a context summary $s_x$:
\[
s_x = A_C(x) = M_\theta(p_C(x))
\]
where $p_C(x)$ is a prompt function designed to elicit clinically relevant contextual information. 

\textcolor{black}{The prompt instructs the LLM to analyze:
glycemic trend: overall trajectory (rising, falling, or stable), rate of change and variability (gradual versus rapid changes, oscillations, or stability) and morphological risk cues: turning points, rebound, plateau patterns, and qualitative hypo-/hyperglycemia risk. Our prompt template is structured as follows:}

\begin{tcolorbox}[colback=yellow!8,
                  colframe=black!50,
                  boxrule=1pt,
                  sharp corners]
\fontsize{7}{9}\selectfont
\textbf{System Role:}  \textit{You are a medical assistant specializing in diabetes management and continuous glucose monitoring. Your task is to summarize the observed CGM history for forecasting support. Use only the information provided in the prompt. Do not use external knowledge, do not assume meal or insulin information, and do not invent unobserved events. Use qualitative descriptions rather than exact numerical future values.}

\vspace{0.3cm}
\textbf{User Prompt:} 
\textit{Analyze the following continuous glucose monitoring readings sampled every 5 minutes over the past 3 hours.}


\vspace{0.2cm}
\textbf{Historical CGM values:} \quad $x_1|x_2|x_3|...|x_{36}$

\vspace{0.2cm}
\textbf{Output Requirements:} \textit{Write a concise clinical-style summary in \textbf{\textit{no more than 5 sentences}}. Describing overall glucose trend, stability or variability, any turning point, rebound, or plateau pattern, and qualitative hypo-/hyperglycemia risk. Do not generate exact future glucose values.}


\end{tcolorbox}

The generated text summaries $s_x$ provide auxiliary signals that complement raw numerical data, capturing qualitative patterns that may be overlooked by purely data-driven models.

\begin{figure*}[h]
\centering
\includegraphics[width=1\linewidth,trim={25 55 25 25},clip]{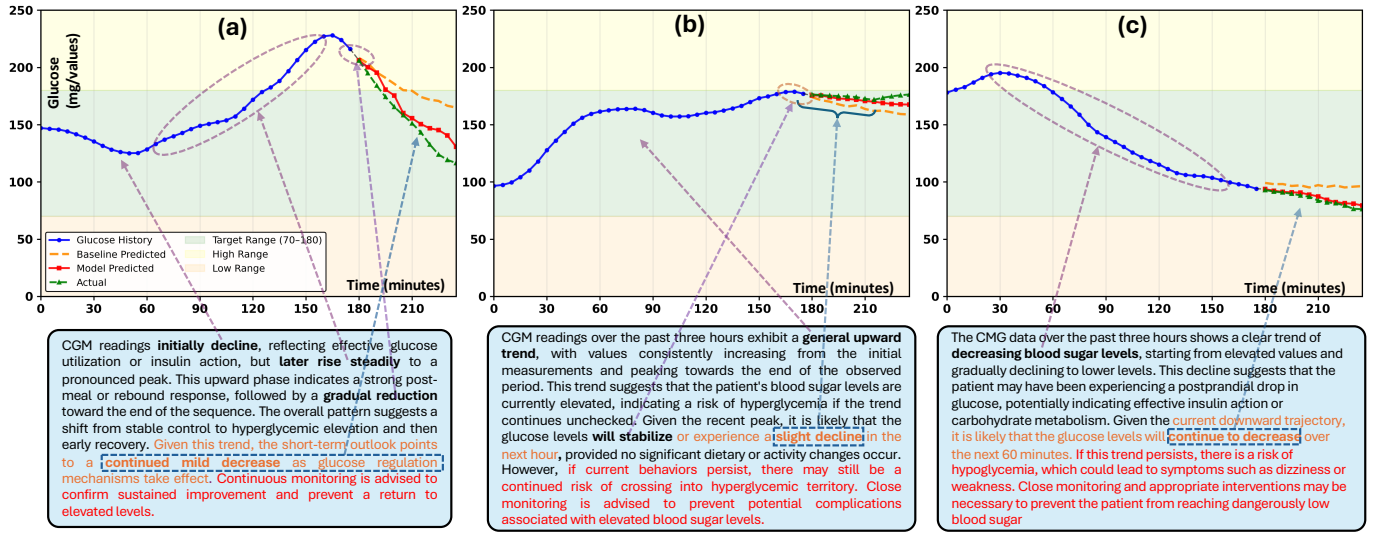}
\vspace{-5mm}
\caption{\color{black}Qualitative effect of contextual summaries on glucose forecasting (three examples).
Panels (a–c) show a 3-hour CGM window (blue), 12-step/60-min forecasts from GlyRAG (\textcolor{red}{red}) and a baseline model (\textcolor{orange}{orange}), and the ground truth future trajectory (\textcolor{green}{green}). Shaded bands indicate clinical ranges (low, target 70–180 mg/dL, high). The callout under each panel is the LLM-generated context summarizing morphology (e.g., rapid rise, rebound, early recovery or sustained decline). GlyRAG leverages this context to anticipate turning points and slope changes, closely tracking the subsequent decrease or stabilization, whereas the baseline tends to overshoot or miss reversals. These examples illustrate how contextual reasoning improves longer-horizon, physiologically coherent forecasts.}
\label{fig:context}
\end{figure*}
\begin{figure*}[t]
\centering
\includegraphics[width=1\linewidth,trim={7 175 5 155},clip]{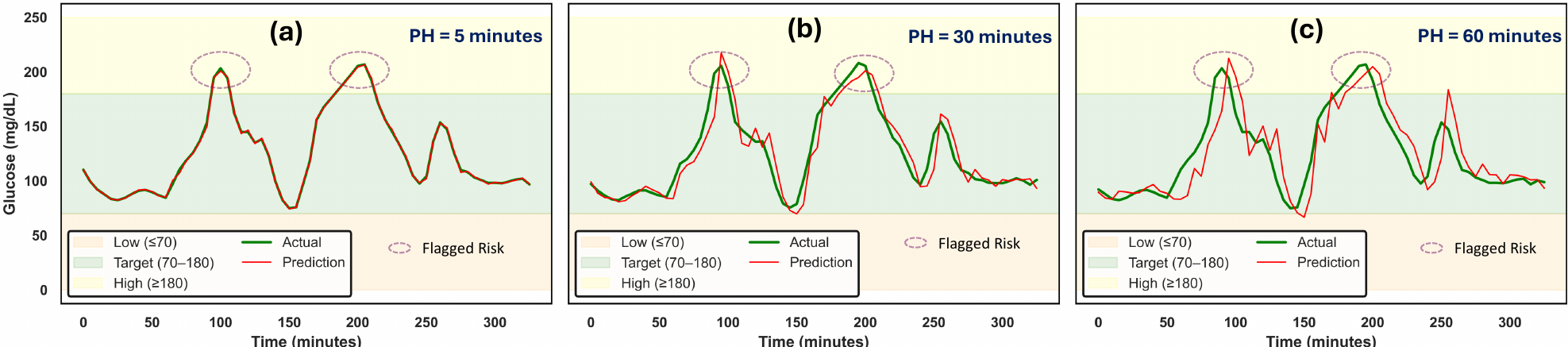}
\vspace{-6.5mm}
\caption{GlyRAG glucose forecasts across prediction horizons (PH = 5, 30, 60 minutes).
GlyRAG predictions (\textcolor{red}{red}) closely follow actual CGM traces (\textcolor{green}{green}) within shaded clinical zones, leveraging contextual morphology to anticipate peaks and nadirs. \textcolor{purple}{Dashed oval} marks risk markers (hypo/hyper) where GlyRAG preserves turning-point fidelity even as the horizon lengthens.}
\label{fig:forecast}
\end{figure*}
\subsection{Multimodal Transformer (MMT) Encoder}
The multimodal transformer encoder integrates contextual information derived from textual summaries with glucose time-series representations. It operates in three stages: context embedding, time-series patch embedding, and cross-modal fusion.

\subsubsection{Context Embedding}
The textual summary $s_x$ is processed through a pre-trained language model (LM) to generate a contextual representation. We employ BERT~\cite{Devlin2019BERTPO} as the context encoder:
\[
z_{\text{text}} = \mathrm{LM}(s_x) \in \mathbb{R}^{d'}
\]
where we extract the [CLS] token embedding from the final layer. This representation is then projected into the multimodal embedding space:
\[
\hat{z}_{\text{context}} = z_{\text{text}} W_{\text{text}} \in \mathbb{R}^{d}
\]
where \( W_{\text{text}} \in \mathbb{R}^{d' \times d} \) is a learnable linear projection with \( d = 512 \).

\subsubsection{Time-Series Patch Embedding}
To capture local temporal patterns, the CGM input sequence 
\( x \in \mathbb{R}^{L} \) is segmented into non-overlapping patches, following the design principles of PatchTST~\cite{PatchTST}. 
Specifically, the sequence is divided into \( N \) patches of length 
\( L_p \) with stride \( L_s \): 
\[
x = [{x}^{(1)}, {x}^{(2)}, \ldots, {x}^{(N)}] 
\in \mathbb{R}^{N \times L_p}, 
\qquad 
N = \left\lceil \frac{L - L_p}{L_s} \right\rceil + 1
\]
For our experiments, we use $L_p =4$, corresponding to 60 minutes at 5-minute sampling, with $L_s =L_p$ . Each patch is then projected into the latent space:
\[
z_{\text{bgl}}^{(i)} = x^{(i)} W_{\text{bgl}}, 
\qquad W_{\text{bgl}} \in \mathbb{R}^{L_p \times d}
\]

The $N$ patch embeddings are aggregated into a single glucose representation:
\[
\hat{z}_{\text{bgl}} = \frac{1}{N} \sum_{i=1}^{N} \hat{z}^{\,i}_{\text{bgl}} 
\in \mathbb{R}^{d}.
\]

\subsubsection{Multi-Modal Fusion via Self-Attention}
Finally, the contextual embedding \( \hat{z}_{\text{context}} \) and the glucose 
embedding \( \hat{z}_{\text{bgl}} \) are fused using multi-head self-attention (MHSA). 
MHSA allows the model to weight contributions dynamically based on the joint signal, preserving modality identity while exchanging information. This enables the fused representation to capture long-range trends (e.g., slow drifts, postprandial arcs) and short-term variations (e.g., rapid drops), while reducing directional bias and stabilizing training.
We first construct a combined sequence representation:
\[
Z = \big[ \hat{z}_{\text{bgl}} \, ; \, \hat{z}_{\text{context}} \big] 
\in \mathbb{R}^{2 \times d}.
\]
For each attention head \( h \in \{1, \ldots, H\} \), the query, key, and value 
matrices are computed as
\begin{equation}
Q_{h} = Z W^{}_{Q_h}, 
\qquad K_{h} = Z W^{}_{K_h}, 
\qquad V_{h} = Z W^{}_{V_h}.
\end{equation}
with \( W^{}_{Q_h}, W^{}_{K_h}, W^{}_{V_h} \in \mathbb{R}^{d \times \frac{d}{H}} \). 
The attention operation is then defined as
\begin{equation}
a^{h} = \text{softmax}\!\left( 
    \frac{Q_{h} K_{h}^{\top}}{\sqrt{d / H}} 
\right) V_{h}
\end{equation}
Outputs from all $H$ heads are concatenated and projected:
\[
Z_{\text{fused}} = \big[ a^{1} \, ; \, a^{2} \, ; \, \ldots \, ; \, a^{H} \big] W^{O} 
\in \mathbb{R}^{2 \times d}.
\]
where $W^{O} \in \mathbb{R}^{d\times d}$ and H=4. 
The fused output is then pooled (e.g., mean over the two tokens) to obtain the multimodal representation
$Z_{\text{fused}}\in\mathbb{R}^{d}$
This fused embedding $Z_{\text{fused}}$ captures bidirectional interactions, enabling $\hat{z}_{\text{bgl}}$ to attend to $\hat{z}_{\text{context}}$, and vice versa. To further enforce consistency between the two modalities, we introduce a cross-translational loss that explicitly aligns contextual and physiological embeddings (described in Section~\ref{sub-sub:ctl}).

\subsubsection{Cross-Translational Loss}
\label{sub-sub:ctl}
To ensure that embeddings from both modalities capture complementary information while maintaining semantic alignment, we introduce a cross-translational loss. This loss encourages 
representations from one modality to be linearly projected into the space of the other, reducing modality mismatch.

Formally, let \( E_{\text{bgl}} \) and \( E_{\text{ctx}} \) denote the glucose 
and context embeddings, respectively. For each modality 
\( k \in \{\text{bgl}, \text{ctx}\} \), and the corresponding paired modality 
\( t \neq k \), we define:
\begin{equation}
\mathcal{L}_{\text{trans}} = 
\sum_{k \in \{\text{bgl}, \text{ctx}\}} 
\sum_{\substack{t \in \{\text{bgl}, \text{ctx}\} \\ t \neq k}} 
\big\| \, \text{Proj}_{k \to t}(E_k) - E_t \, \big\|_2^2    
\end{equation}
where $\text{Proj}_{k \to t}$ is a learnable linear projection from modality $k$ to modality $t$. This formulation ensures that the contextual and glucose signals remain mutually informative, thereby stabilizing multimodal fusion and enhancing forecasting generalization. It also mitigates modality collapse, preventing the model from disregarding one source of information.

\subsection{Retrieval-Augmented (RAG) Forecasting}
\subsubsection{Neighbor Retrieval}
Once the multimodal encoder is pretrained, we construct an embedding database 
from the training set. For each training sample \((x_j, y_j)\), the fused 
representation is computed as
\[
z_j = \text{Encoder}(x_j, s_{x_j}) \in \mathbb{R}^{d},
\]
where \( s_{x_j} \) denotes the LLM-generated context corresponding to the input 
sequence \( x_j \). The collection of all embeddings paired with their outcomes 
forms the retrieval database
\[
\mathcal{D} = \{ (z_j, y_j) : j = 1, \ldots, |\text{Train}| \}.
\]

During inference, a test input window \( x_{\text{test}} \) is 
passed through the MMT to produce an embedding \( Z_{\text{test}} \). Using cosine 
similarity, the system retrieves the top-\(K\) most similar embeddings from the 
training set:
\begin{equation}
\mathcal{N}(Z_{\text{test}}) =\{(Z_{jk},y_{jk}):j_k \in \operatorname*{arg\,top-K}_{j\in \mathcal{D}} 
\frac{Z_{\text{test}} \cdot Z_{j}}
{\|Z_{\text{test}}\| \, \|Z_{j}\|}\} 
\end{equation}
\subsubsection{Cross-attention adapter (query–neighbor fusion)}
We then fused $Z_\text{test}$ with each neighbor via cross-attention branches. For branch $i \in {1\cdots K}$ with $m_H$ heads:
\[
H^{(i)} =
\text{Concat}\!\left(
\text{Attn}\!\big(
Z_{\text{test}} W_{Q_h}^{(i)},
\, z_{j_i} W_{K_h}^{(i)},
\, z_{j_i} W_{V_h}^{(i)}
\big)
\right)_{h=1}^{m_H} W_H^{(i)},
\]

where the attention operation is defined as
\[
\text{Attn}(Q,K,V) =
\text{Softmax}\!\left(\frac{QK^{\top}}{\sqrt{d}}\right) V,
\]
with learnable projection matrices
\(
W_{Q_h}^{(i)}, W_{K_h}^{(i)}, W_{V_h}^{(i)} \in \mathbb{R}^{d \times d}
\)
and
\(
W_H^{(i)} \in \mathbb{R}^{(m_H d) \times d}
\).

The branch outputs are aggregated (e.g., mean or learned weights) to produce a retrieval-aware representation:
\[
z_{\text{rag}} = \text{Aggregate}\!\big( H^{(1)}, \ldots, H^{(K)} \big)
\in \mathbb{R}^{d}.
\]
The forecast is produced by an MLP head over the query and retrieval features (transformer frozen; only the adapter/MLP is fine-tuned):
\vspace{-2mm}
\[
\vspace{-2mm}
\hat{y} = f_{\text{MLP}}\!\left(
    [ \, Z_{\text{test}} \, ; \, z_{\text{rag}}\ ]
\right)
\]

In our experiments we set $K=3$. 
The retrieval mechanism enables the model to adapt predictions based on similar historical patterns, effectively implementing a form of case-based reasoning that leverages the learned embedding space.

\begin{algorithm}[t]
\caption{\small 
GlyRAG for BGL forecasting (Inference). \small
\(x^{\text{test}}_{1:L} \in \mathbb{R}^{L}\): input window and \(H\): forecast horizon. 
\(A_C(\cdot)\): contextualization agent; 
\(E_{\text{LM}}\): text encoder;
\(W_{\text{ctx}}\): projection to \(d\);
\(E_{\text{bgl}}\): CGM encoder to \(d\);
\(F_{\omega}\): two-token fusion (MHSA) with params \(\omega\);
\(A_{\psi}\): retrieval adapter (neighbor cross-attention + mixing) with params \(\psi\);
\(f_{\phi} : \mathbb{R}^{2d} \!\to\! \mathbb{R}^{H}\): forecast head with params \(\phi\). \\[2pt]
\(\mathcal{D} = \{ (z_j, y_j) \}_{j=1}^{n}\): retrieval index storing fused representations and targets from training;
\(K\): number of neighbors;
}
\label{alg:glyrag-inference}
\textbf{Input:} \ \(x^{\text{test}}_{1:L}\); retrieval index \(\mathcal{D}\); configuration \((H, K)\)\\
\textbf{Output:} Forecast \(\hat{y} \in \mathbb{R}^{H}\)
\begin{algorithmic}[1]
\small
\State \textbf{Begin}
\State \( s_x \leftarrow M_{\theta}(p(x^{\text{test}}_{1:L}))\)
\State \(\tilde{z}_{\text{ctx}} \leftarrow E_{\text{LM}}(s_x) W_{\text{ctx}}\) 
        \hfill\{morphology → context vector\}
\State \( \bar{z}_{\text{bgl}} \leftarrow E_{\text{bgl}}(x^{\text{test}}_{1:L}) \)
        \hfill\{Encode CGM to \(d\)-dim\}
\State \( z \leftarrow F_{\omega}([\bar{z}_{\text{bgl}}; \tilde{z}_{\text{ctx}}]) \)
        \hfill\{Fuse CGM + context\}
\State \( \mathcal{N}(z) \leftarrow \arg\!\top_K 
        \frac{z^\top z_j}{\|z\|_2 \|z_j\|_2} \text{ from } \mathcal{D} \)
        \hfill\{\(K\) nearest neighbors\}
\State \( z_{\text{rag}} \leftarrow 
        A_{\psi}(z, \{z_j\}_{j \in \mathcal{N}(z)}) \)
        \hfill\{Cross-attn + mixing;\}
\State \( \hat{y} \leftarrow f_{\phi}([z; z_{\text{rag}}]) \)
        \hfill\{Forecast \(H\) future points\}
\State \( \hat{y} \leftarrow \text{Denorm}(\hat{y}) \)
        \hfill\{Denormalization\}
\State \Return \( \hat{y} \)
\State \textbf{End}
\end{algorithmic}
\end{algorithm}

\section{Experimental Setup}
\subsection{Datasets}
We evaluated the proposed framework using two large-scale, real-world T1D datasets that capture diverse physiological and behavioral patterns under both conventional and automated insulin delivery settings. 

We chose the OhioT1DM~\cite{marling2020ohiot1dm} to demonstrate the results of our proposed methods. 
\textcolor{black}{Although the dataset is relatively small, containing only 12 subjects, it still offers notable diversity. It includes balanced gender representation (6 male, 6 female), a broad age range (20–80 years), and two cohorts collected in different releases: the 2018 version with Basis Peak bands and the 2020 version with Empatica Embrace bands. These characteristics make the OhioT1DM dataset a valuable benchmark, despite its modest size, and explain its frequent adoption in the literature for blood glucose prediction research.}


To assess generalization to contemporary automated insulin delivery (AID) use, we also analyze AZT1D~\cite{AZT1D}, a new clinic-sourced dataset collected at Mayo Clinic (Scottsdale, AZ) between Dec 2023–Apr 2024. The cohort includes 25 adults (13 female, 12 male; age 27–80, mean 59), each contributing on average ~26 days of real-world data. AZT1D contains Dexcom G6 Pro CGM (5-min sampling), Tandem t:slim X2 pump logs (including granular bolus details: total dose, bolus type, correction components), carbohydrate intake, and device mode (regular/sleep/exercise). In total, the release comprises 320,488 CGM readings spanning ~26,707 hours. Compared with OhioT1DM, AZT1D offers richer therapy context under AID, enabling studies on morphology-aware forecasting, decision support, and patient-twin personalization. 
\subsection{\color{black}Preprocessing and Leakage Prevention}
\begin{figure}[]
\vspace{-5mm}
\centering
\includegraphics[width=1\linewidth,trim={5 5 5 4},clip]{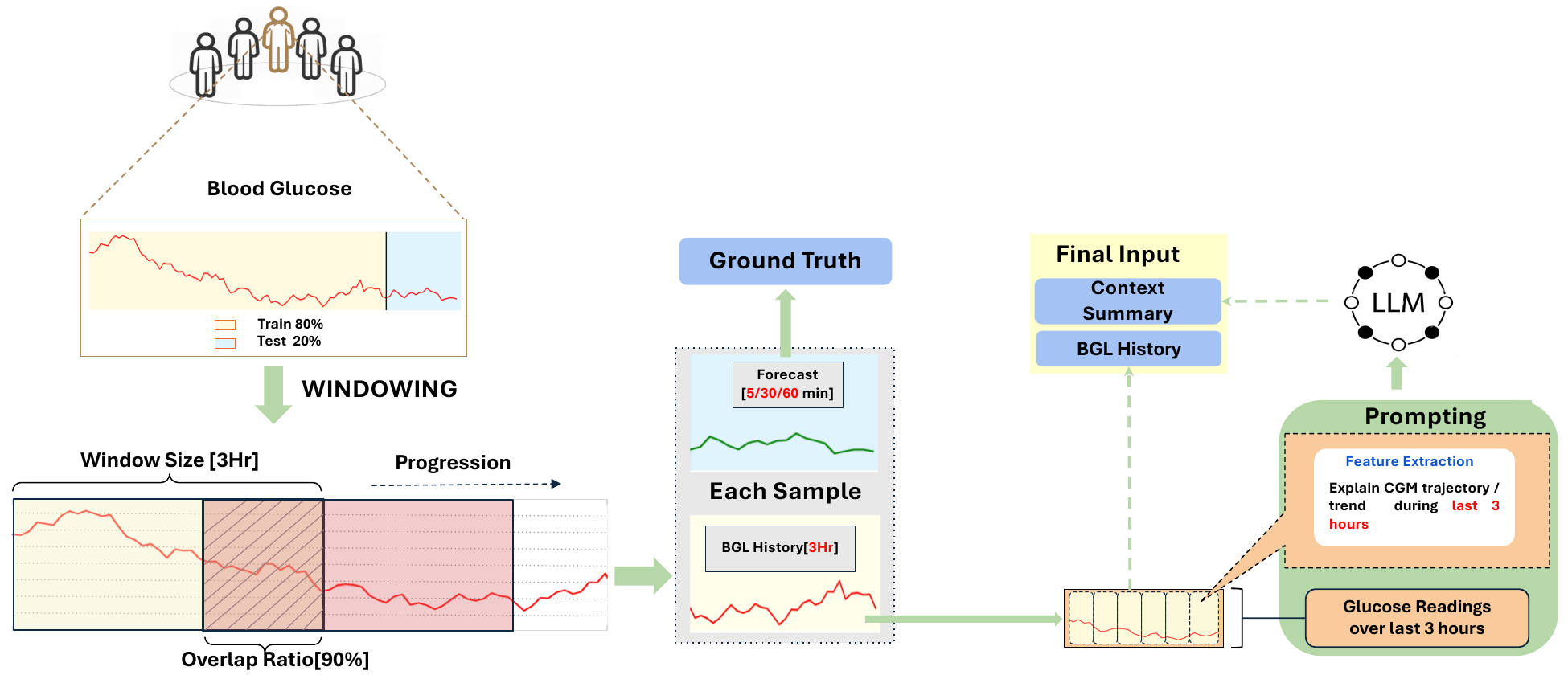}
\vspace{-5mm}
\caption{\textbf{Preprocessing and context‑extraction workflow}: CGM streams are segmented into overlapping 3‑h windows with 5/30/60‑min targets; an LLM summarizes each window into a context token that is concatenated with the 3‑h glucose history to form GlyRAG’s input.}
\vspace{-7mm}
\label{fig:preprocess}
\end{figure}
The dataset is divided into training and testing datasets for each participant. The models are trained on the training dataset. Prediction metrics are computed using the test dataset for each participant. The model's input consists of a three-hour (180 minutes) sliding window of
historical data, providing sufficient information for making
good predictions for the next prediction horizon (PH). Three different
PHs (5, 30 and 60 minutes) are considered for comparison.

Each patient’s raw CGM sequence is first segmented into fixed-length windows of length $L$, corresponding to the input horizon for forecasting. For our main experiments, we use $L=36$, which represents a three-hours history of glucose values. Each input window is paired with prediction targets at 5-, 30-, and 60-minute horizons. 

\textcolor{black}{
To prevent train--test and temporal leakage, all participant records are split before window construction. We use the dataset-provided OhioT1DM split and a chronological split for AZT1D, then generate sliding windows separately within each split so that no input window or prediction target crosses the train--test boundary. Normalization statistics are fit only on each participant's training split and applied unchanged to validation/test windows. The retrieval index is also built only from same-participant training windows; test windows are used only as queries and are never inserted into the retrieval memory. In the primary personalized experiments, GlyRAG does not retrieve from other participants. We use $K=3$ nearest neighbors in the main setting, selected from the sensitivity analysis in Section~\ref{subsec:context_retrieval_ablation}. 
} 
\textcolor{black}{
All experiments use personalized models trained and evaluated separately for each participant. OhioT1DM uses the provided split ($\sim$44 training days and $\sim$12 test days per participant), while AZT1D uses a chronological participant-wise split before windowing. We did not estimate a universal minimum training duration, since this depends on participant-specific variability; all baselines and GlyRAG use identical splits for fair comparison.
}

\subsection{Training Objective}
The overall training objective is a weighted combination of the forecasting 
loss and the cross-translational loss:
\[
\mathcal{L} = \mathcal{L}_{\text{forecast}} + \lambda \mathcal{L}_{\text{trans}}
\]
where the hyperparameter \(\lambda\) balances predictive accuracy with modality 
alignment. This formulation ensures that the model not only forecasts accurately 
but also learns meaningful relationships between glucose trajectories and 
contextual summaries.

\begin{table*}[t]
\vspace{-6mm}
\caption{\color{black}Comparison of GlyRAG with state-of-the-art blood glucose forecasting models. The best values are colored in red. BGL: Blood Glucose, I: Insulin, C: Carb.}
    \centering
    \setlength{\tabcolsep}{2.5pt} 
    \footnotesize
    \begin{tabular}{cc|c|*{2}{c}|*{2}{c}|*{2}{c}}
    \toprule
        \multicolumn{3}{c|}{\textbf{Prediction Horizon (PH)}} &\multicolumn{2}{c|}{\textbf{\makecell{5 min \\Next Sample}}} & \multicolumn{2}{c|}{\textbf{30 min}} & \multicolumn{2}{c}{\textbf{60 min}}\\
        \hline
        Dataset & \textbf{Study} & \textbf{Modalities} & RMSE & MAE & RMSE & MAE & RMSE & MAE \\
         
         \midrule
         \multirow{11}{*}{\centering Ohio} 
         & CNN-RNN~\cite{daniels2021multitask} & BGL, I, C, E &- &- & 18.8 & 13.2 & 31.8 & 23.4\\
         & MTL-LSTM~\cite{shuvo2023deep} & BGL, I, C &- &- &16.06 & 10.64 & 30.89 & 22.07\\
         & GlySim~\cite{arefeen2023glysim} & BGL, I, C &11.3 &8.2 & 17.5 & 12.3 & 24.2 & 16.5 \\

         & \textcolor{black}{GLIMMER-CNN-LSTM}~\cite{glimmer} & BGL &
\textcolor{black}{17.12} & \textcolor{black}{13.62} &
\textcolor{black}{26.01} & \textcolor{black}{19.93} &
\textcolor{black}{37.82} & \textcolor{black}{28.73} \\

& \textcolor{black}{GLIMMER-Transformer}~\cite{glimmer} & BGL &
\textcolor{black}{30.58} & \textcolor{black}{28.35} &
\textcolor{black}{35.50} & \textcolor{black}{31.02} &
\textcolor{black}{43.67} & \textcolor{black}{36.39} \\

         &TimesFM~\cite{das2024decoder}    &BGL  &3.22 &1.96 &11.46 &6.69 &21.71 &13.05\\
        \cmidrule{2-9}
        & \textcolor{black}{PatchTST$^\dagger$} & BGL &
\textcolor{black}{4.29} & \textcolor{black}{2.65} &
\textcolor{black}{13.78} & \textcolor{black}{8.71} &
\textcolor{black}{23.11} & \textcolor{black}{14.98} \\

& \textcolor{black}{iTransformer$^\dagger$} & BGL &
\textcolor{black}{4.42} & \textcolor{black}{2.76} &
\textcolor{black}{13.85} & \textcolor{black}{8.72} &
\textcolor{black}{23.22} & \textcolor{black}{15.03} \\

& \textcolor{black}{Chronos (zero-shot)$^\dagger$} & BGL &
\textcolor{black}{4.49} & \textcolor{black}{2.82} &
\textcolor{black}{14.92} & \textcolor{black}{9.30} &
\textcolor{black}{25.28} & \textcolor{black}{16.17} \\
        
        \cmidrule{2-9}
        &PatchTST + GlyRAG (Ours) &BGL, Context &\textbf{1.97}$^*$ &\textbf{1.83}$^*$ &\textbf{10.61}$^*$ &\textbf{6.19}$^*$ &\textbf{20.22}$^*$ &\textbf{12.33}$^*$\\
        & \textcolor{black}{95\% CI} & \textcolor{black}{--} &
\textcolor{black}{[1.92, 2.02]} & \textcolor{black}{[1.72, 1.94]} &
\textcolor{black}{[10.48, 10.74]} & \textcolor{black}{[6.08, 6.34]} &
\textcolor{black}{[20.1, 20.35]} & \textcolor{black}{[12.34, 12.45]} \\
        \bottomrule
    \end{tabular}
    
    \vspace{1mm}
    \begin{minipage}{\linewidth}
    \footnotesize
    \color{black}
    $\dagger$ BGL-only baseline reproduced in this work without context and retrieval component. 
    * $p<0.05$.
    \end{minipage}
    \label{tab:sota-performance-compare}
\end{table*}

\begin{table}[]
\vspace{-5mm}
\color{black}
\centering
\caption{\color{black}Long-horizon RMSE against strong CGM-only baselines.}
\label{tab:main_rmse_corrected}
\footnotesize
\setlength{\tabcolsep}{2pt}
\renewcommand{\arraystretch}{1}
\begin{tabular}{c|cc|cc}
\toprule
\multirow{2}{*}{\textbf{Model}} &
\multicolumn{2}{c|}{\textbf{Ohio}} &
\multicolumn{2}{c}{\textbf{AZT1D}} \\
\cmidrule(lr){2-3}\cmidrule(lr){4-5}
& \textbf{30} & \textbf{60} & \textbf{30} & \textbf{60} \\
\midrule
\textbf{GPT-4 GlyRAG}
& \makecell{\textbf{10.61}$^*$\\{[10.48,10.74]}}
& \makecell{\textbf{20.22}$^*$\\{[20.09,20.35]}}
& \makecell{\textbf{10.11}$^*$\\{[9.98,10.24]}}
& \makecell{\textbf{16.10}$^*$\\{[15.97,16.23]}} \\
\hline
LLaMA GlyRAG
& \makecell{11.63$^*$\\{[11.30,12.00]}}
& \makecell{22.02$^*$\\{[21.70,22.34]}}
& \makecell{13.06$^*$\\{[12.85,13.27]}}
& \makecell{20.47$^*$\\{[20.26,20.68]}} \\
\midrule
PatchTST
& 13.78 & 23.11 & 13.42 & 22.18 \\
iTransformer
& 13.85 & 23.22 & 13.77 & 22.30 \\
Chronos
& 14.92 & 25.28 & 15.25 & 25.50 \\
\bottomrule
\end{tabular}

\vspace{1mm}
\begin{minipage}{0.98\linewidth}
    \footnotesize
    \color{black}Values in brackets are 95\% CIs. $^*$ Significant improvement over PatchTST using participant-level Wilcoxon signed-rank test ($p<0.05$).
\end{minipage}
\vspace{-2mm}
\end{table}

\subsection{Evaluation Metrics}
To comprehensively assess forecasting performance, we employ four complementary evaluation metrics: root mean square error (RMSE), mean absolute error (MAE), and Clarke Error Grid Analysis (CEGA) 
and Pearson correlation coefficient ($r$). 
These metrics jointly capture numerical accuracy, clinical relevance, and correlation strength between predicted and actual glucose values.

\subsubsection{Root Mean Square Error (RMSE)}The RMSE measures the average magnitude of prediction errors with a higher penalty on large deviations, providing insight into worst-case predictive performance:
\[
\text{RMSE} = \sqrt{ \frac{1}{n} \sum_{i=1}^{n} \big( y_i - \hat{y}_i \big)^{2} }
\]
where $n$ is the number of test samples, $y_i$ denotes the ground-truth CGM value, and $\hat{y_i}$  is the predicted value.

\subsubsection{Mean Absolute Error (MAE)} The MAE captures the average absolute deviation between predicted and observed glucose values, offering a straightforward interpretation of prediction accuracy:
\[
\text{MAE} = \frac{1}{n} \sum_{i=1}^{n} \big| y_i - \hat{y}_i \big|
\]

\subsubsection{Clarke Error Grid (CEGA)}
To evaluate the clinical relevance of predictions, we use Clarke Error Grid Analysis~\cite{clarkGrid}, which classifies predicted glucose values into five zones (A–E) based on their proximity to reference values. Zone A represents clinically accurate predictions, Zone B represents benign errors, while Zones C–E indicate increasingly dangerous misclassifications that could adversely affect treatment decisions. CEGA is widely used in diabetes research to assess the safety of predictive algorithms.

\subsubsection{Pearson Correlation Coefficient ($r$)}
The Pearson correlation quantifies the linear relationship between predicted 
and actual glucose values:
\[
r = \frac{\sum_{i=1}^{n} (y_i - \bar{y}) \, (\hat{y}_i - \bar{\hat{y}})}
{\sqrt{\sum_{i=1}^{n} (y_i - \bar{y})^2} \; \sqrt{\sum_{i=1}^{n} (\hat{y}_i - \bar{\hat{y}})^2}}
\]
where \(\bar{y}\) and \(\bar{\hat{y}}\) are the sample means of the actual and 
predicted glucose values, respectively.

\subsection{Architecture Configuration}
\textcolor{black}{
CGM windows are patch-embedded with patch length $6$ and stride $3$, combined with sinusoidal positional encoding, and processed by a Transformer encoder ($d_{\mathrm{model}}=512$, $n_{\mathrm{layers}}=3$, $n_{\mathrm{heads}}=4$, $d_{\mathrm{ff}}=2048$, $\mathrm{dropout}=0.05$). A pretrained BERT-family language model produces a $768$-dimensional text/context embedding, which is projected to $d_{\mathrm{model}}=512$ and appended as a context token. The encoder outputs are reshaped per variable and summarized by a two-layer LSTM forecast head with hidden size $256$ for multi-horizon prediction.
}

\textcolor{black}{During pretraining, we use a bidirectional cross-modal translator with three-layer MLPs and hidden size $512$. The objective combines Huber forecasting loss and weighted translation loss ($\alpha=0.1$). The Huber transition parameter is fixed at the default PyTorch value ($\delta=1.0$) for all experiments and is not tuned on the validation set. Models are trained with a batch size of $64$ using AdamW with a learning rate of $3\times10^{-4}$ and weight decay of $10^{-4}$.
}

\textcolor{black}{After pretraining, the backbone is frozen and training-set encoder embeddings are indexed for cosine-similarity retrieval. The retrieval module uses $k=3$ nearest neighbors unless otherwise specified. Retrieved analogues are processed by a compact cross-attention module and a small LSTM. The retrieval hidden state is concatenated with the pooled forecast embedding and passed to a final MLP prediction head.
}

\textcolor{black}{
All context summaries were generated offline before model training and cached for reuse. For GPT-4, we used deterministic decoding with temperature $=0$. For LLaMA-3.1, we used greedy decoding with temperature $=0$, \texttt{do\_sample=false}, and a maximum output length of 120 tokens. During forecasting, GlyRAG uses the cached context embedding and retrieval adapter; the LLM is not queried online. All experiments were conducted on an A100 GPU.
}

\section{Results}
We evaluated the proposed GlyRAG framework against representative state-of-the-art blood glucose forecasting models \textcolor{black}{and recent time-series foundation models, including TimesFM and Chronos,} on two benchmark datasets across multiple prediction horizons. All methods were re-implemented or evaluated following their original experimental settings, and performance was assessed using standard forecasting error metrics as well as clinically motivated measures reported in subsequent tables.

\subsection{Compare with State-of-the-Art (SOTA)}


~\tblref{sota-performance-compare} summarizes the quantitative comparison with representative blood glucose forecasting methods at 5-, 30-, and 60-minute prediction horizons. \textcolor{black}{The compared methods include recurrent, multitask, transformer-based, and foundation-model baselines, with input modalities ranging from CGM-only to CGM plus insulin (I)/carbohydrate (C) records. In contrast, the main GlyRAG configuration uses only CGM-derived context, allowing a modality-matched comparison with CGM-only baselines while avoiding dependence on additional behavioral or therapy inputs.}

\textcolor{black}{
Overall, GlyRAG achieves the lowest RMSE and MAE across the evaluated horizons on the full OhioT1DM cohort. The largest gains appear at clinically useful longer horizons, where recent CGM continuity alone becomes less sufficient and morphology-aware context can help the model anticipate trend changes, plateaus, and rebounds. Compared with the strongest reproduced CGM-only baseline, PatchTST, GlyRAG reduces RMSE from 13.78 to 10.61 at 30 minutes and from 23.11 to 20.22 at 60 minutes. This improvement suggests that the benefit is not only due to the PatchTST backbone, but also to the LLM-derived context and retrieval mechanisms, which are discussed more in Section~\ref{subsec:context_retrieval_ablation}.
}

\textcolor{black}{
Table~\ref{tab:main_rmse_corrected} further reports the long-horizon RMSE comparison against strong CGM-only baselines, including 95\% confidence intervals and participant-level Wilcoxon signed-rank tests for statistical significance. GPT-4 with GlyRAG significantly outperforms PatchTST at both horizons across both datasets ($p<0.05$). LLaMA+GlyRAG shows smaller but still significant long-horizon gains, indicating that the contextualization pipeline is not limited to GPT-4. We focus this compact statistical table on longer horizons (30 \& 60 minutes) because these horizons provide more actionable lead time for diabetes decision support than next-sample prediction.}

\begin{figure}[]
\centering
\includegraphics[width=1.02\linewidth,trim={16 10 45 12},clip]{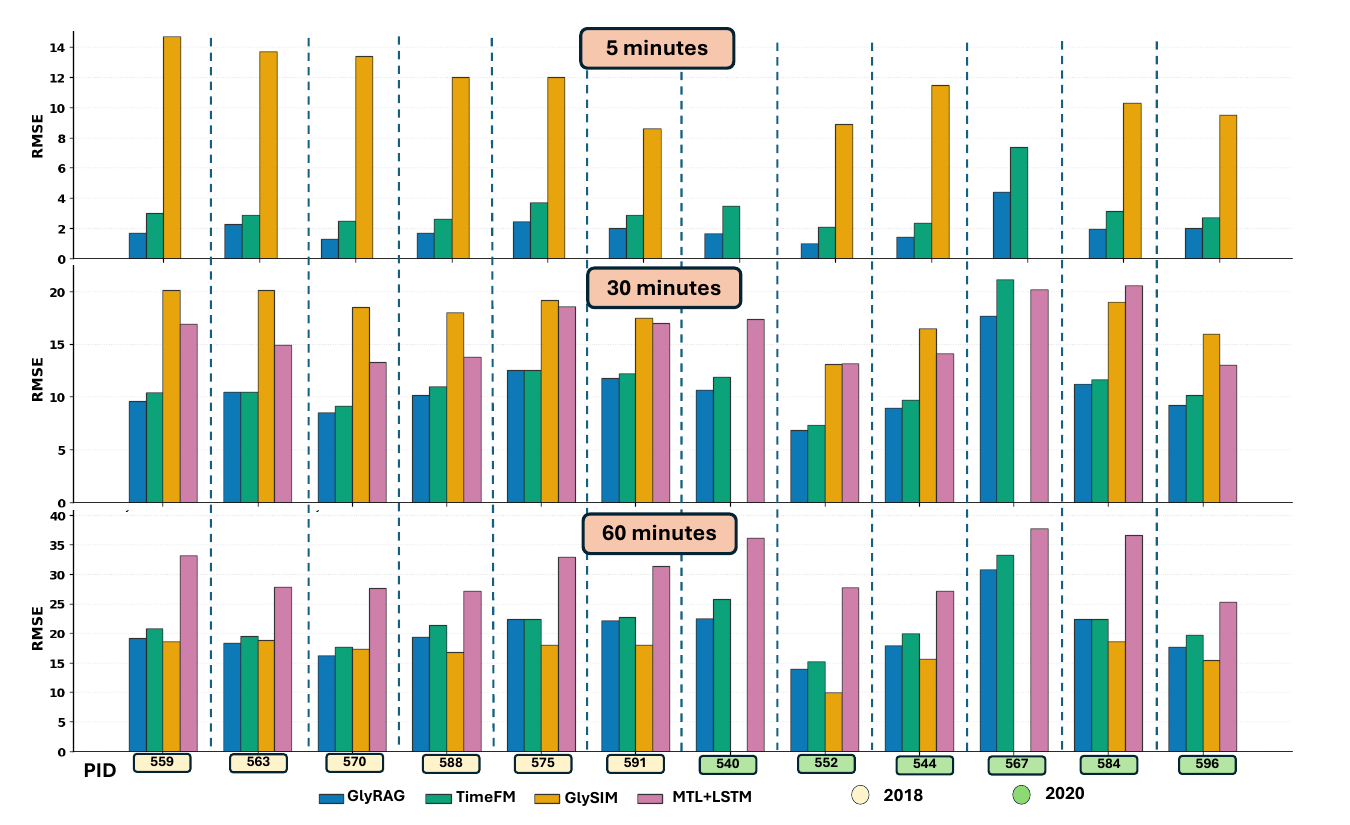}
\vspace{-4mm}
\caption{Patient-wise RMSE Comparison Across Prediction Horizons for GlyRAG and Baselines.}
\vspace{-4mm}
\label{fig:patient_wise_rmse}
\end{figure}

\subsection{Clinical Evaluation}
Accurate clinical evaluation of glucose forecasting models requires metrics that capture both numerical precision and medical relevance. Sensitivity analysis measures the model’s ability to detect critical dysglycemic events—hypoglycemia and hyperglycemia—where early intervention is vital for patient safety. The Clarke Error Grid (CEG) and Continuous Glucose–Error Grid Analysis (CG‑EGA) are widely used clinical frameworks that assess how prediction errors may translate into treatment risks. While CEG evaluates the clinical safety of predicted glucose values across zones A–E, CG‑EGA jointly considers point and rate‑of‑change errors to examine the physiological coherence of temporal trends. Complementary to these, Time‑in‑Range (TIR) quantifies how well predicted glucose levels align with the clinically optimal range (typically 70–180 mg/dL), serving as an aggregate measure of overall glycemic stability. Together, these metrics provide a comprehensive evaluation of forecasting reliability, safety, and clinical utility in diabetes management.

\begin{table}[!h]
\vspace{-5mm}
\caption{Clinical evaluation at PH = 60 minutes.
Hyper-/hypoglycemia sensitivity and CEG distributions (Zones A–E). GlyRAG concentrates predictions in Zones A–B with competitive sensitivity and shows smaller TIR deviation than the baseline, indicating clinically reliable forecasting.}
\centering
\footnotesize
\setlength{\tabcolsep}{2.5pt}

\begin{tabular}{
>{\columncolor{gray!8}}c|
>{\columncolor{blue!6}}l
>{\columncolor{blue!10}}l|
>{\columncolor{green!6}}c
>{\columncolor{green!8}}c
>{\columncolor{green!10}}c
>{\columncolor{green!12}}c
>{\columncolor{green!14}}c|
>{\columncolor{red!10}}c
}
\toprule
\multicolumn{9}{c}{\textbf{\textit{Ohio Dataset, PH = 60 minutes}}}\\
\midrule
\multirow{2}{*}{\textbf{Method}} &
\multicolumn{2}{c|}{\cellcolor{blue!15}\textbf{Sensitivity}} &
\multicolumn{5}{c}{\cellcolor{green!20}\textbf{Clarke Error Grid Regions (\%)}} & P-Coeff \\
\cmidrule(lr){2-3}\cmidrule(lr){4-8}
& Hyper & Hypo & A$\uparrow$ & B$\downarrow$ & C$\downarrow$ & D$\downarrow$ & E$\downarrow$ &\textit{(r)} \\
\midrule
\textcolor{black}{GLIMMER (T)} &
\textcolor{black}{91} &
\textcolor{black}{0.0} &
\textcolor{black}{47.41} &
\textcolor{black}{49.18} &
\textcolor{black}{\textbf{0.03}} &
\textcolor{black}{3.30} &
\textcolor{black}{0.07} &
\textcolor{black}{0.787}\\

\textcolor{black}{GLIMMER (CL)} &
\textcolor{black}{81} &
\textcolor{black}{15} &
\textcolor{black}{62.88} &
\textcolor{black}{33.49} &
\textcolor{black}{0.75} &
\textcolor{black}{2.87} &
\textcolor{black}{0.02} &
\textcolor{black}{0.784}\\
TimesFM~\cite{das2024decoder} &92&81&75.95&22.36&0.29&1.33&0.07&0.918\\
\midrule
\rowcolor{yellow!10}
\textbf{GlyRAG} &
\textbf{97} &
\textbf{92} &
\textbf{85.53} &
\textbf{13.59} &
0.15 &
\textbf{0.83} &
\textbf{0.025} &
\textbf{0.942}\\
\bottomrule
\end{tabular}

\vspace{2mm}

\begin{tabular}{
>{\columncolor{gray!8}}c|
>{\columncolor{blue!6}}c
>{\columncolor{blue!10}}c|
>{\columncolor{green!6}}c
>{\columncolor{green!8}}c
>{\columncolor{green!10}}c
>{\columncolor{green!12}}c
>{\columncolor{green!14}}c|
>{\columncolor{red!10}}c
}
\toprule
\multicolumn{8}{c}{\textbf{\textit{AZT1D Dataset, PH = 60 minutes}}}\\
\midrule
\multirow{2}{*}{\textbf{Method}} &
\multicolumn{2}{c|}{\cellcolor{blue!15}\textbf{Sensitivity}} &
\multicolumn{5}{c}{\cellcolor{green!20}\textbf{Clarke Error Grid Regions (\%)}} & P-Coeff\\
\cmidrule(lr){2-3}\cmidrule(lr){4-8}
& Hyper & Hypo & A$\uparrow$ & B$\downarrow$ & C$\downarrow$ & D$\downarrow$ & E$\downarrow$ &\textit{(r)}\\
\midrule
\textcolor{black}{GLIMMER (T)} &
\textcolor{black}{71.89} &
\textcolor{black}{0.00} &
\textcolor{black}{54.22} &
\textcolor{black}{44.24} &
\textcolor{black}{0.00} &
\textcolor{black}{1.38} &
\textcolor{black}{0.15} &
\textcolor{black}{0.653}\\

\textcolor{black}{GLIMMER (CL)} &
\textcolor{black}{63.60} &
\textcolor{black}{3.91} &
\textcolor{black}{64.75} &
\textcolor{black}{33.48} &
\textcolor{black}{0.17} &
\textcolor{black}{1.57} &
\textcolor{black}{0.03} &
\textcolor{black}{0.652}\\
TimesFM~\cite{das2024decoder} &88.2&\textbf{60.4}&69.94&28.43&0.26&1.27&0.09&0.821\\
\midrule
\rowcolor{yellow!10}
\textbf{GlyRAG} &
\textbf{92} &
44.2 &
\textbf{84.9} &
\textbf{13.6} &
\textbf{0.04} &
\textbf{0.38} &
\textbf{0.02} &
\textbf{0.84}\\
\bottomrule
\end{tabular}

\vspace{2mm}

\begin{tabular}{C{2cm}|C{2cm}C{2cm}}
\toprule
\rowcolor{cyan!10}
\multicolumn{3}{c}{\textbf{\textit{Time In Range (TIR) difference between observed and predicted}}}\\
\midrule
\textbf{Method} & \textbf{Ohio} & \textbf{AZT1D}\\
\midrule
\textbf{GlyRAG} & \textbf{0.91$\pm$0.85} & \textbf{0.70$\pm$0.60}\\
Baseline &1.27$\pm$0.96 &0.97$\pm$0.78\\
\bottomrule
\end{tabular}
\label{tab:clinical_metrics}
\vspace{-3mm}
\end{table}

\begin{table*}[!t]
\centering
\small
\vspace{-6mm}
\caption{CG-EGA results at PH = 60 min for the Ohio and AZT1D cohorts, stratified by hypoglycemic, euglycemic, and hyperglycemic ranges. Metrics report Accurate Prediction (AP), Benign Error (BE), and Erroneous Prediction (EP) rates; higher AP and lower EP indicate improved clinical safety. \textcolor{red}{Red: best values in each category.} \textcolor{black}{For each dataset, the first row reports GlyRAG and the parenthesized second row reports the BGL-only baseline.}}
\label{tab:patient_cgega_compact}
\footnotesize
\setlength{\tabcolsep}{1.5pt}
\renewcommand{\arraystretch}{1.2}
\footnotesize
\begin{tabular}{
c|
ccc|
ccc|
ccc|
ccc
}
\toprule
\multirow{2}{*}{\rotatebox[origin=c]{75}{\textbf{ \fontsize{7}{9}\selectfont Dataset}}} &
\multicolumn{3}{c|}{\textbf{Hypo ($\leq$70mg/dL)}} &
\multicolumn{3}{c|}{\textbf{Eu (70-180 mg/dL)}} &
\multicolumn{3}{c|}{\textbf{Hyper ($\geq$180 mg/dL)}} &
\multicolumn{3}{c}{\textbf{Average}}\\
& AP$\uparrow$ & BE & EP$\downarrow$ & AP$\uparrow$ & BE & EP$\downarrow$ & AP$\uparrow$ & BE & EP$\downarrow$& AP$\uparrow$ & BE & EP$\downarrow$\\
\midrule
\multirow{2}{*}{\rotatebox[origin=c]{75}{\textbf{Ohio}}}
& \color{red}92.5$\pm$5.3  & \color{red}5.1$\pm$4.2  & \color{red}2.3$\pm$2.1
& \color{red}94.6$\pm$4.1  & \color{red}5.0$\pm$3.4 & 0.42$\pm$0.7
& \color{red}92.2$\pm$6.9  & 7.3$\pm$5.8 & \color{red}0.7$\pm$1.0
& \color{red}93.0$\pm$5.4 & \color{red}5.8$\pm$4.5 & \color{red}1.1$\pm$1.3 \\
& (88.4$\pm$18) & (7.06$\pm$9.2) & (4.5$\pm$9.3)
& (94.4$\pm$3.9) & (5.2$\pm$3.3)  & \color{red}(0.39$\pm$0.6)
& (92.0$\pm$6.9) & \color{red}(7.2$\pm$5.8) & \color{red}(0.7$\pm$1.2)
& (91.6$\pm$9.6) & (6.5$\pm$6.1)  & (1.9$\pm$3.7) \\
\midrule
\multirow{2}{*}{\rotatebox[origin=c]{75}{\textbf{ \fontsize{6.5}{9}\selectfont AZT1D}}}
& 65.4$\pm$22.9  & 12.3$\pm$10.4 & \color{red}19.0$\pm$18.7
& 82.4$\pm$9.7   & 16.1$\pm$8.6  & 1.5$\pm$0.9
& \color{red}71.6$\pm$18.0  & \color{red}25.1$\pm$15.6 & \color{red}3.4$\pm$2.9
& \color{red}73.1$\pm$16.9  & \color{red}17.8$\pm$11.5 & \color{red}7.9$\pm$7.5 \\
& \color{red}(65.7$\pm$23.2)  & \color{red}(11.7$\pm$9.9)  & (22.53$\pm$20.7)
& (81.6$\pm$9.6)   & (16.9$\pm$8.8)  & (1.55$\pm$1)
& (70.7$\pm$18.1)  & (25.9$\pm$16.1) & \color{red}(3.4$\pm$2.6)
& (72.7$\pm$17)  & (18.7$\pm$11.8) & (9.16$\pm$8.1) \\
\bottomrule
\end{tabular}
\label{tab:cg_ega_summary}
\end{table*}
\subsubsection{Clarke Error Grid and TIR Analysis}
~\tblref{clinical_metrics} summarizes clinical performance at PH = 60 min using clarke error grid (CEG), event sensitivities, pearson correlation ($r$), and time-in-range (TIR) deviation for the Ohio and AZT1D cohorts. GlyRAG concentrates most predictions in clinically acceptable CEG Zones A–B and maintains low proportions in the clinically dangerous Zones D–E, indicating a reduced likelihood of forecasts that could lead to inappropriate treatment actions.

GlyRAG also shows balanced sensitivity to hypo- and hyperglycemic events—especially hypoglycemia, where timely detection is critical—while preserving specificity and temporal agreement (higher $r$) with observed traces. For example, in the Ohio cohort, GlyRAG achieves 14\% improvement in hypoglycemia sensitivity compared with the best prior model, TimesFM. The comparatively lower hypoglycemia sensitivity observed in AZT1D is attributable to the limited number of hypoglycemic events in that cohort, a known challenge for event-based evaluation that reduces statistical power rather than indicating systematic model failure. We also compared GlyRAG to TimesFM-a foundation time-series model that uses glucose trajectories alone and found that explicitly modeling CGM-derived context yields superior clinical reliability. 

The lower mean absolute TIR deviation ($<$1) versus the baseline further confirms that GlyRAG better preserves overall glycemic exposure over the forecast horizon. \textcolor{black}{Together, these results indicate that GlyRAG improves long-horizon forecasting while maintaining favorable clinical-error profiles.}

\begin{figure}[!h]
\centering
\includegraphics[width=1.5\linewidth,trim={182 77 20 80},clip]{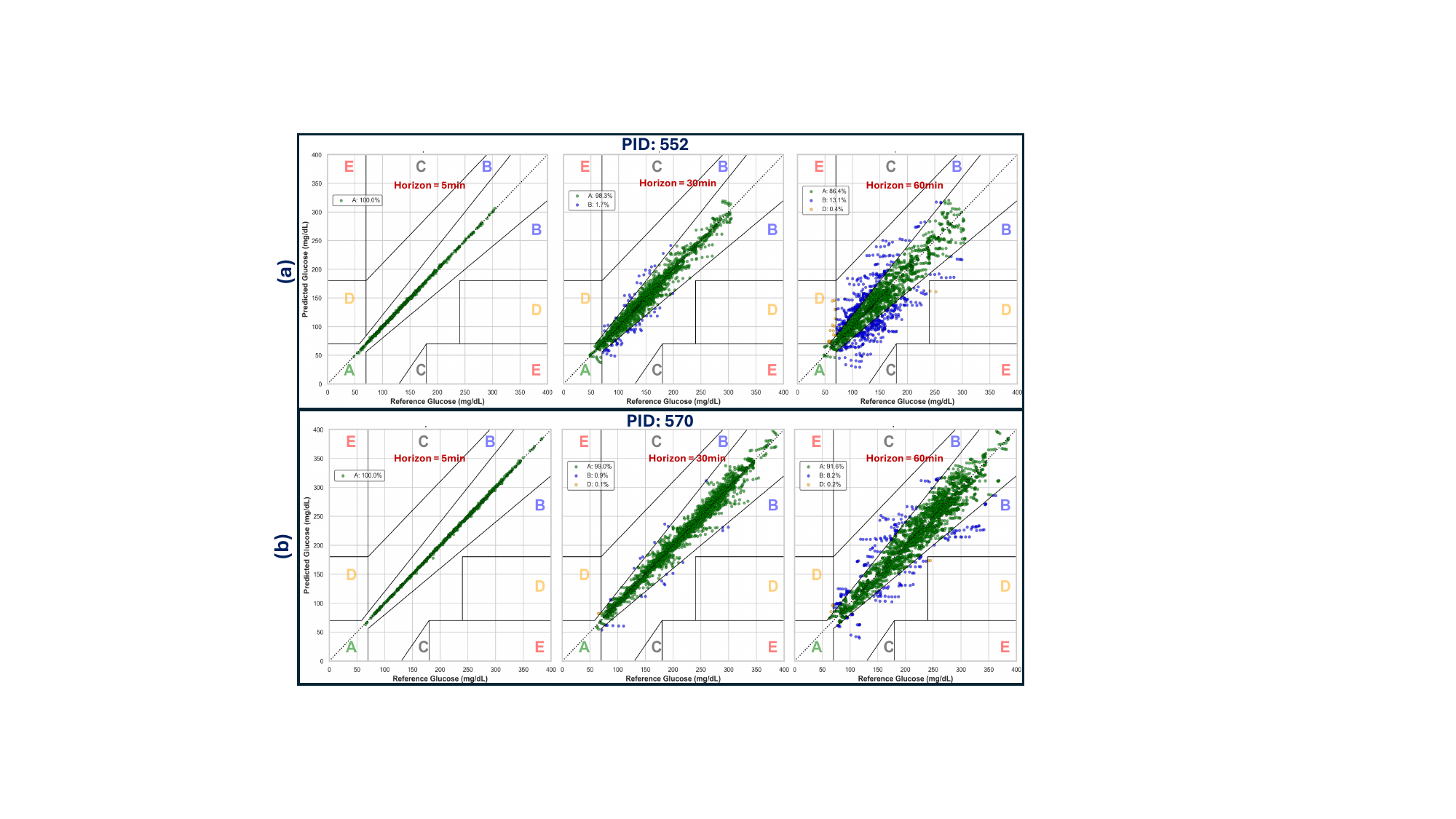}
\vspace{-6.5mm}
\caption{Clarke Error Grid analysis for Patients 552 (a) and 570 (b) across 5, 30 and 60-minute prediction horizons. Most predictions fall within Zone A, indicating high clinical accuracy, with minor dispersion into Zone B at longer horizons, showing slightly reduced but reliable forecasting performance.}
\label{fig:cg_error}
\end{figure}

Overall, GlyRAG shows lower mean absolute TIR deviation than the baseline and concentrates predictions in safe CEG Zones A–B, indicating better preservation of overall glycemic exposure and fewer clinically risky forecasts. These results underscore that context augmentation materially improves long-horizon CGM forecasting for decision support.

\begin{figure}[]
\centering
\includegraphics[width=1\linewidth,trim={5 6 8 310},clip]{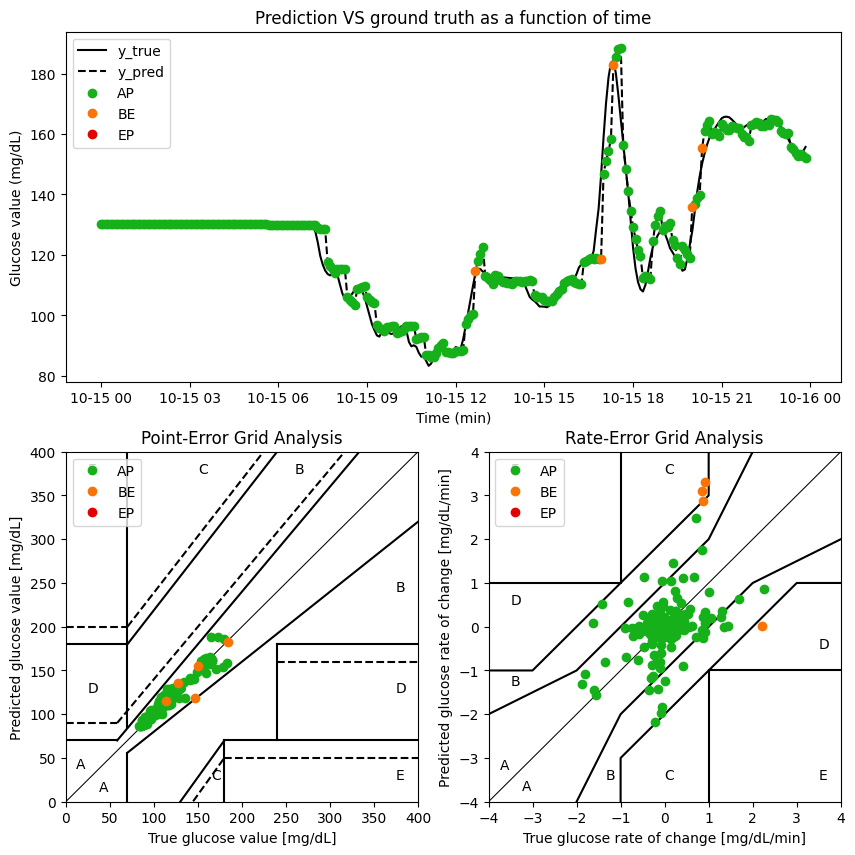}
\vspace{-2mm}
\caption{Patient‑level CG‑EGA (point‑ and rate‑error grids) for an Ohio participant (PID 512).}
\vspace{-4mm}
\label{fig:cg_ega_ohio}
\end{figure}

\subsubsection{Continuous Glucose–Error Grid Analysis (CG-EGA)}~\tblref{patient_cgega_compact} reports the CG-EGA at 60-minute PH, stratified by hypoglycemic, euglycemic, and hyperglycemic ranges for both datasets. GlyRAG achieves high Accurate Prediction (AP) rates across glucose ranges, particularly in the euglycemic region, while maintaining low Erroneous Prediction (EP) percentages; on average, the overall EP decreases from 1.9\% to 1.1\% on Ohio and from 9.2\% to 7.9\% on AZT1D, indicating fewer clinically unsafe predictions. This pattern indicates that the model preserves clinically acceptable trend and point accuracy without increasing the risk of misleading predictions during physiologically stable periods. 
As illustrated in ~\figref{cg_ega_ohio}, the patient‑level CG‑EGA for a representative Ohio subject at PH = 60 min shows that most forecasts (green AP points) fall in Zones A/B of both the point and rate‑error grids, with only a few BE/EP points, indicating clinically acceptable accuracy in both glucose values and rates of change.

Importantly, GlyRAG maintains competitive AP and controlled EP rates in hypoglycemic and hyperglycemic ranges, which are most relevant for safety-critical decision making. Although variability increases in extreme glucose ranges—consistent with prior CGM forecasting studies—the EP values remain limited, suggesting that contextual modeling helps mitigate clinically unsafe trend errors. These CG-EGA results complement the Clarke Error Grid (CEGA) findings in ~\tblref{clinical_metrics}, where predictions are concentrated in Zones A–B with strong event sensitivity and reduced TIR deviation.

\begin{figure}[!b]
\centering
\includegraphics[width=1\linewidth,trim={15 20 15 20},clip]{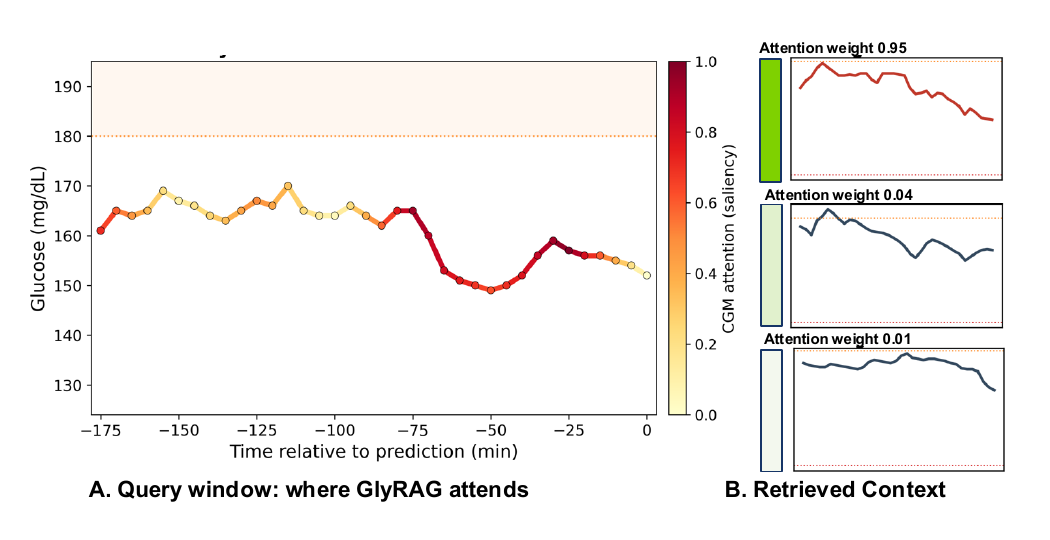}
\vspace{-5mm}
\caption{\color{black}\textbf{Morphology-aware retrieval example.} 
Panel A shows the query CGM window with retrieval-adapter saliency. Panel B shows the top retrieved training episodes from the same participant with learned attention weights. GlyRAG assigns the highest weight to the most morphologically similar episode, indicating selective case-based retrieval rather than simple averaging.}
\vspace{-3mm}
\label{fig:attention}
\end{figure}

\subsection{\textcolor{black}{Context and Retrieval Ablation}}
\label{subsec:context_retrieval_ablation}

\begin{table*}[]
\color{black}
\centering
\caption{Context and retrieval ablation. Values report percentage RMSE improvement over the PatchTST CGM-only baseline $B$; values in parentheses show the corresponding no-RAG setting using the same context. Positive values indicate lower RMSE than the baseline.}
\label{tab:context_retrieval_ablation}
\footnotesize
\setlength{\tabcolsep}{3.2pt}
\renewcommand{\arraystretch}{1.12}
\begin{tabular}{ll|ccc|ccc}
\toprule
\multirow{2}{*}{LLM/Agent} & \multirow{2}{*}{Context Strategy} &
\multicolumn{3}{c|}{OhioT1DM} &
\multicolumn{3}{c}{AZT1D} \\
\cmidrule(lr){3-5}\cmidrule(lr){6-8}
& & 5 min & 30 min & 60 min & 5 min & 30 min & 60 min \\
\midrule
\multirow{5}{*}{GPT-4}
& $B0$: rule-based CGM morphology
& -1.5 (0.0) & -0.3 (1.3) & -1.9 (1.1)
& 0.3 (-0.1) & 0.0 (1.1) & 1.3 (1.6) \\
& $B1$: rule-based + CGM statistics
& 0.6 (0.7) & -4.0 (0.3) & -1.1 (1.0)
& 0.0 (0.4) & -1.2 (1.5) & 4.3 (1.2) \\
& $M0$: LLM narrative, CGM-only
& \textbf{54.1} (30.4) & \textbf{23.0} (16.7) & \textbf{12.5} (11.0)
& \textbf{25.6} (20.1) & \textbf{24.7} (21.9) & \textbf{27.4} (22.8) \\
& $M1$: LLM narrative + CGM statistics
& 34.1 (19.0) & 14.6 (12.6) & 9.3 (10.0)
& 12.2 (10.2) & 21.6 (20.0) & 25.6 (23.7) \\
& $M2$: LLM narrative + events
& 40.7 (29.2) & 16.1 (10.0) & 9.5 (10.2)
& 25.1 (18.4) & 22.5 (21.2) & 26.3 (23.2) \\
\midrule
\multirow{5}{*}{LLaMA-3.1}
& $B0$: rule-based CGM morphology
& 0.0 (-1.3) & -2.3 (2.1) & -2.3 (2.1)
& 0.6 (-0.6) & -1.0 (1.7) & 2.8 (5.5) \\
& $B1$: rule-based + CGM statistics
& 1.6 (0.6) & -3.3 (2.3) & -3.3 (2.3)
& 0.4 (0.5) & -1.6 (2.1) & 6.3 (2.2) \\
& $M0$: LLM narrative, CGM-only
& \textbf{1.9} (1.2) & 15.6 (13.9) & \textbf{4.7} (13.3)
& -1.2 (-0.2) & \textbf{2.7} (11.4) & \textbf{7.7} (13.0) \\
& $M1$: LLM narrative + CGM statistics
& -0.2 (0.4) & 14.6 (14.1) & 4.6 (12.1)
& -0.2 (0.2) & 1.6 (10.8) & 5.6 (13.1) \\
& $M2$: LLM narrative + events
& 0.7 (1.1) & \textbf{16.1} (14.9) & 4.1 (13.9)
& -0.1 (0.1) & 2.5 (11.4) & 6.3 (13.2) \\
\bottomrule
\end{tabular}
\end{table*}

\begin{figure*}[t]
\vspace{-4mm}
\centering
\subfloat[\color{black}OhioT1DM.]{
    \includegraphics[width=0.48\linewidth,trim={5 10 5 40},clip]{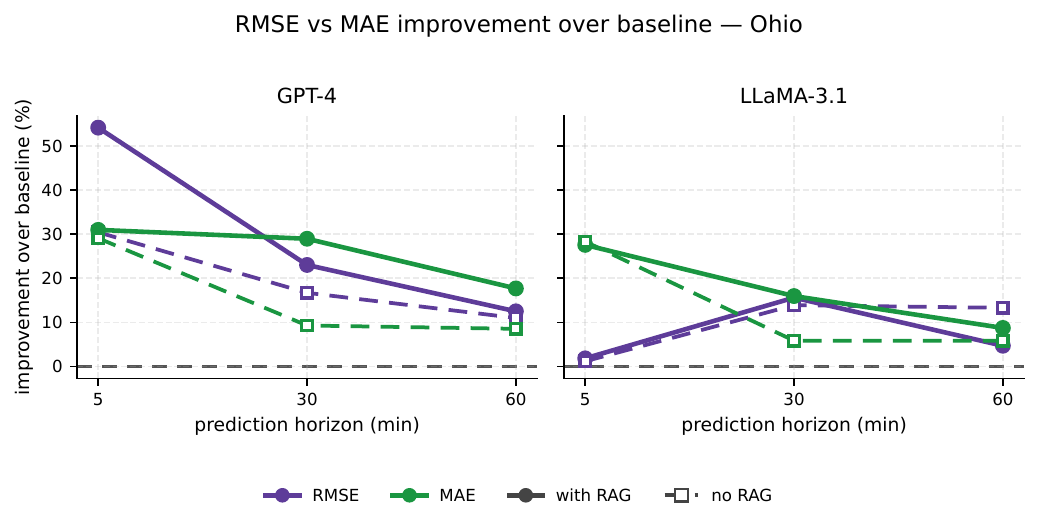}
    \label{fig:rag_curve_ohio}
}
\hfill
\subfloat[\color{black}AZT1D.]{
    \includegraphics[width=0.48\linewidth,trim={5 10 5 40},clip]{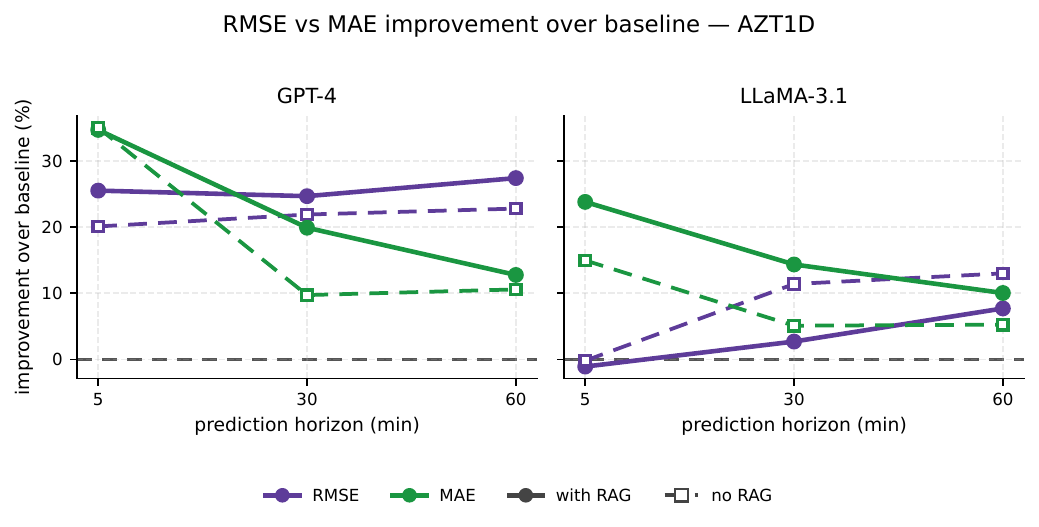}
    \label{fig:rag_curve_azt1d}
}

\vspace{-4mm}

\subfloat[\color{black}Isolated RAG effect.]{
    \includegraphics[width=0.8\linewidth,trim={8 8 10 40},clip]{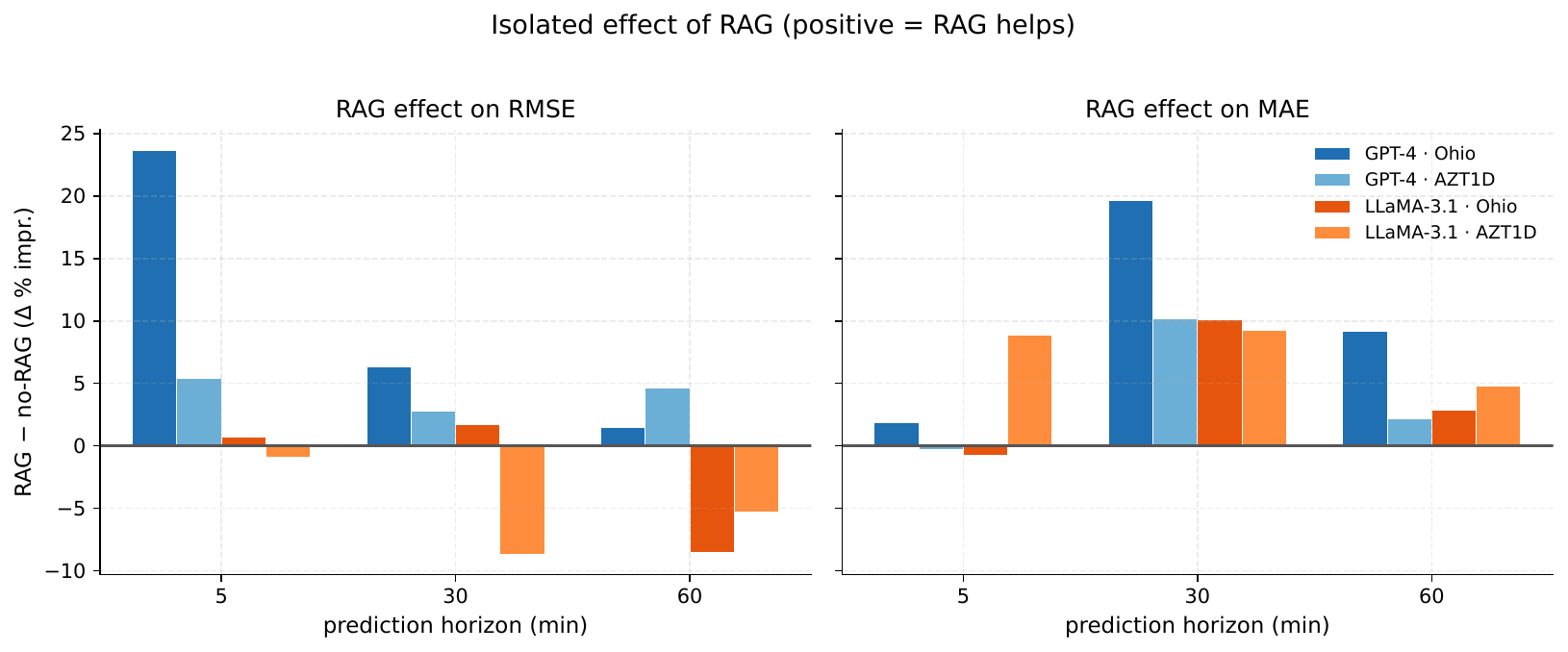}
    \label{fig:rag_isolated_effect}
}

\vspace{-2mm}
\caption{\color{black}\textbf{Effect of LLM context and retrieval.}
(a)--(b) RMSE/MAE improvement over the CGM-only baseline with and without RAG. 
(c) Marginal retrieval effect computed as RAG minus no-RAG improvement; positive values indicate that retrieval helps.}
\vspace{-3mm}
\label{fig:rag_effect_summary}
\end{figure*}

\textcolor{black}{
To separate the effects of context generation and retrieval,~\tblref{context_retrieval_ablation} compares GlyRAG variants with and without RAG against the PatchTST CGM-only baseline $B$. Here, $B0$ and $B1$ denote non-LLM rule-based context variants, $M0$ is the main CGM-only LLM narrative setting, $M1$ adds CGM-derived statistics, and $M2$ adds carbohydrate/insulin event aggregates only as a non-CGM ablation. 
\newline
Fig.~\ref{fig:rag_effect_summary} visualizes the same ablation from Table~\ref{tab:context_retrieval_ablation}. Panels (a) and (b) show that the no-RAG variants already improve over the CGM-only baseline in many settings, indicating that the LLM context token itself contributes substantially. Panel (c) isolates the marginal effect of retrieval by plotting RAG minus no-RAG improvement. Retrieval provides an additional but model-dependent refinement, with the clearest gains for GPT-4, while the effect is smaller and occasionally negative for LLaMA-3.1.
The non-LLM variants ($B0$, $B1$) test whether a handcrafted context branch is sufficient, while the LLM variants test whether morphology-aware summaries add value. The main setting is $M0$, where the LLM receives only the past CGM window; $M1$ adds CGM-derived statistics, and $M2$ adds carbohydrate/insulin events only as a non-CGM ablation.
}
\begin{figure}[t]
\centering
\subfloat[\textcolor{black}{Accuracy sensitivity to retrieval size $K$.}]{
    \includegraphics[width=0.82\linewidth,trim={0 0 0 20},clip]{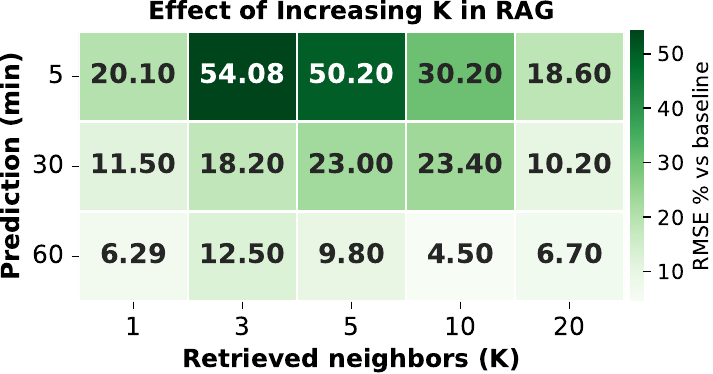}
    \label{fig:k_sensitivity_acc}
}

\vspace{0.5mm}

\subfloat[\textcolor{black}{Latency change relative to $K=1$.}]{
    \includegraphics[width=0.8\linewidth,trim={0 10 0 20},clip]{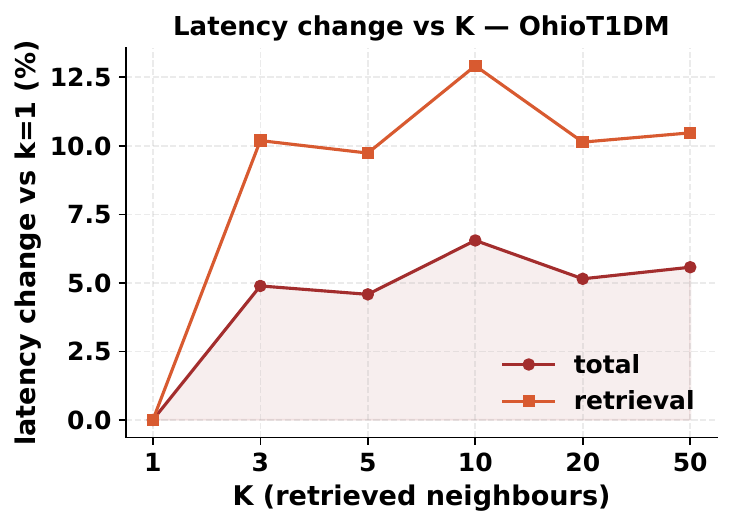}
    \label{fig:k_sensitivity_latency}
}

\vspace{-1.5mm}
\caption{\color{black}\textbf{Effect of retrieval size K.}
(a) RMSE improvement over baseline across retrieval sizes and prediction horizons. 
(b) Relative latency change compared with K=1. 
K=3 gives a practical accuracy–latency trade-off.}
\vspace{-3mm}
\label{fig:k_sensitivity}
\end{figure}

\textcolor{black}{The results show three main trends. First, handcrafted context variants ($B0$, $B1$) do not consistently improve over the CGM-only baseline, with many gains near zero or negative. Second, the CGM-only LLM narrative ($M0$) provides the most reliable gains, especially with GPT-4, reaching 54.1\% RMSE improvement at the 5-min horizon on OhioT1DM and staying above 24\% across all AZT1D horizons. This suggests that the LLM summary captures useful glucose morphology beyond simple rules or deterministic statistics.} Third, comparing values outside and inside parentheses shows that retrieval provides an additional but model-dependent \textcolor{black}{refinement: for GPT-4 on OhioT1DM at 5 min, RAG increases the improvement from 30.4\% to 54.1\%, whereas for LLaMA-3.1 the gain is smaller and sometimes neutral or negative. Fig.~\ref{fig:attention} qualitatively illustrates this behavior: the retrieval adapter assigns most attention to the historical episode whose CGM morphology best matches the query window, supporting the interpretation of RAG as selective case-based refinement rather than simple averaging.}

\textcolor{black}{
We further evaluate sensitivity to the retrieval size $K$ in~\figref{k_sensitivity_acc}. Performance is best or near-best for small retrieval sets, with $K=3$ providing the most stable accuracy--complexity trade-off across horizons. Larger $K$ values tend to dilute the retrieval signal by introducing less similar historical episodes.~\figref{k_sensitivity_latency} shows the latency cost of increasing the retrieval size. Moving from $K=1$ to $K=3$ increases \color{black}\textit{total inference latency} by only about 4.6\%, while \textit{retrieval latency} increases by about 9.6\%. This supports using $K=3$ as a practical accuracy--latency trade-off.}

\begin{figure}[]
\centering
\includegraphics[width=0.95\linewidth,trim={8 8 5 0},clip]{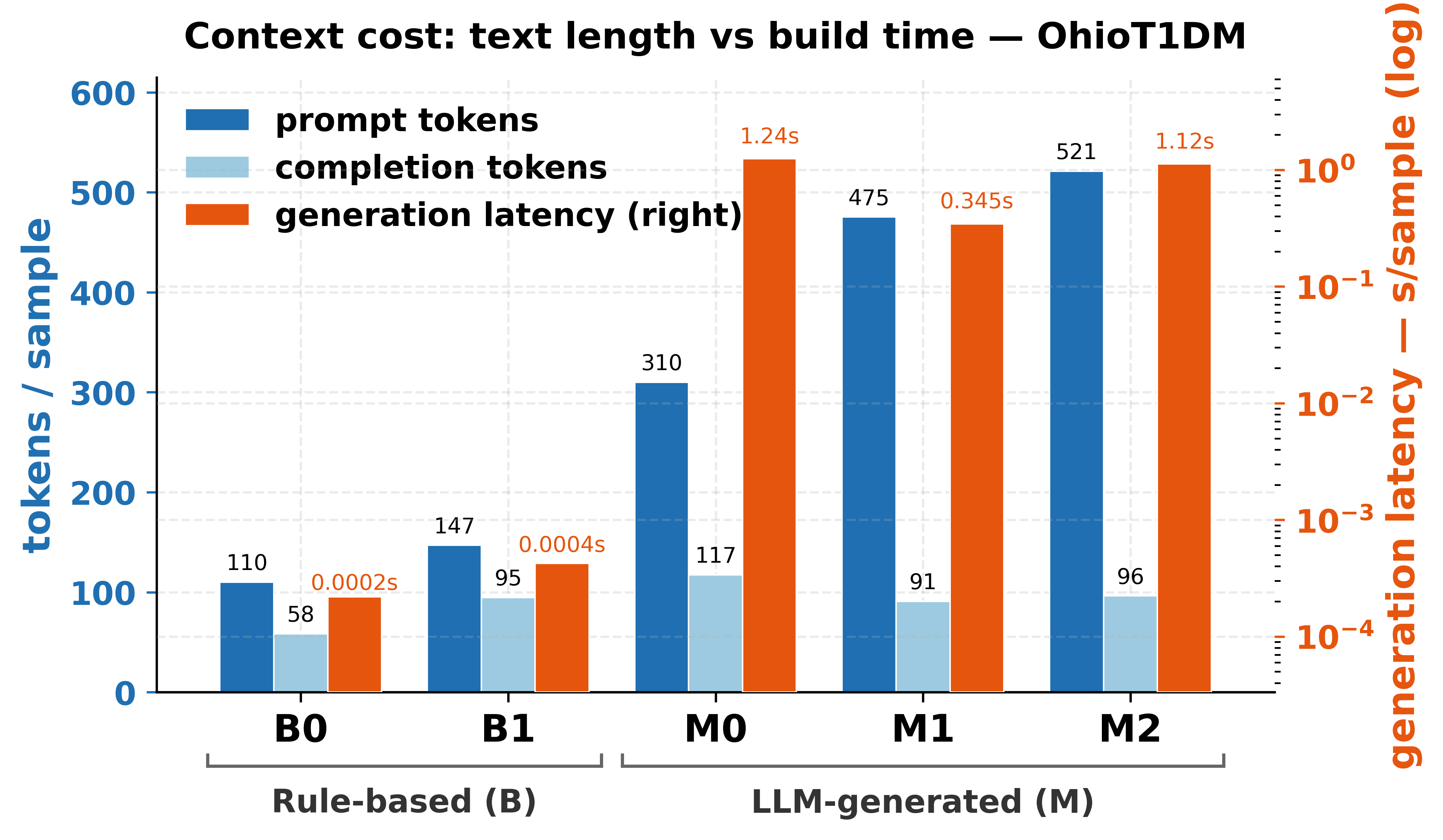}
\vspace{-2mm}
\caption{\color{black}Token counts and generation latency for rule-based and LLM-generated context variants on Ohio dataset.}
\vspace{-3mm}
\label{fig:context_cost}
\end{figure}
\textcolor{black}{
~\figref{context_cost} also compares the cost of constructing each context variant. Rule-based variants are nearly free, whereas LLM-generated variants require offline generation and caching. Among the LLM variants, $M0$ uses the shortest CGM-only prompt and avoids event information, making it the most practical context strategy.
}
\textcolor{black}{Finally, adding more prompt information does not guarantee better performance: neither CGM statistics ($M1$) nor carbohydrate/insulin events ($M2$)} \textcolor{black}{consistently outperform the simpler CGM-only $M0$ setting. Therefore, we use $M0$ as the primary GlyRAG configuration.
}

\subsection{Ablation Study}

\begin{table}[h]
\vspace{-2mm}
\setlength{\tabcolsep}{5.5pt}
\small
\caption{Ablation Study to see the effect of different components in GlyRAG}
\centering
\footnotesize
\begin{tabular}{c|cc|l|>{\columncolor{blue!6}}c
>{\columncolor{blue!10}}c
>{\columncolor{blue!18}}c|
>{\columncolor{green!5}}c
>{\columncolor{green!12}}c
>{\columncolor{green!20}}c} 
\toprule
\multicolumn{10}{c}{\textbf{\textit{Ohio Dataset}}}\\
\midrule

\multirow{2}{*}{\rotatebox[origin=c]{90}{\textbf{RAG}}}& \multicolumn{2}{c|}{\textbf{Context}} 
& \multirow{2}{*}{\textbf{\rotatebox[]{90}{BGL}}} 
& \multicolumn{3}{c|}{\cellcolor{blue!18}\textbf{RMSE}} 
& \multicolumn{3}{c}{\cellcolor{green!18}\textbf{MAE}}\\
\cmidrule(lr){2-3}\cmidrule(lr){5-7}\cmidrule(lr){8-10}
& CA & CTL &  & 5 & 30 & 60 & 5 & 30 & 60 \\
\midrule
     \multirow{3}{*}{\cmark} 
     & \cmark & \cmark & \cmark & \textbf{1.97} & \textbf{10.61} & 20.22 & 1.43 & \textbf{6.19} & 12.33 \\
     & \cmark & \xmark & \cmark & 2.03 & 10.90 & \textbf{20.17}      &1.2      &6.22       &\textbf{12.29} \\  
     & \xmark & \cmark & \cmark & 2.07       &10.84       &20.41       &1.26      &6.43       &12.56 \\
     \midrule
     \xmark & \cmark & \cmark & \cmark &1.98&10.95&20.83 &1.81 &6.25&12.42\\
     \midrule
     \xmark & \xmark & \xmark & \cmark & 2.04  & 11.71 & 21.59 & \textbf{1.18} & 6.31  & 12.65 \\
\bottomrule
\end{tabular}

\vspace{1mm}

\begin{tabular}{c|cc|l|>{\columncolor{blue!6}}c
>{\columncolor{blue!10}}c
>{\columncolor{blue!18}}c|
>{\columncolor{green!5}}c
>{\columncolor{green!12}}c
>{\columncolor{green!20}}c} 
\toprule
\multicolumn{10}{c}{\textbf{\textit{AZT1D Dataset}}}\\
\midrule
\multirow{2}{*}{\rotatebox[origin=c]{90}{\textbf{RAG}}} 
& \multicolumn{2}{c|}{\textbf{Context}} 
& \multirow{2}{*}{\rotatebox[origin=c]{90}{\textbf{BGL}}} 
& \multicolumn{3}{c|}{\cellcolor{blue!18}\textbf{RMSE}} 
& \multicolumn{3}{c}{\cellcolor{green!18}\textbf{MAE}}\\
\cmidrule(lr){2-3}\cmidrule(lr){5-7}\cmidrule(lr){8-10}
& CA & CTL &  & 5 & 30 & 60 & 5 & 30 & 60 \\
\midrule
     \multirow{3}{*}{\cmark}& \cmark & \cmark & \cmark &\textbf{4.17}	&13.52	&\textbf{22.32}	&\textbf{2.98}	&9.48	&\textbf{15.24}\\      
     & \cmark & \xmark & \cmark &4.18	&\textbf{13.45}	&22.46	&3.05	&\textbf{9.11}	&15.31\\
     & \xmark & \cmark & \cmark &4.60	&14.12	&23.01	&3.29	&9.69	&16.37\\
     \midrule
     
     \xmark & \cmark & \cmark & \cmark &4.19 &13.70 &23.07 &3.05 &9.53 &15.46\\
     \midrule
     \xmark & \xmark & \xmark & \cmark &4.22	&13.48	&23.52	&3.04	&9.17	&15.49\\
\bottomrule
\end{tabular}
\vspace{-2mm}
\label{tab:ablation-study}
\end{table}
~\tblref{ablation-study} evaluates the contribution of retrieval (RAG) and contextual components—cross‑attention (CA) and cross‑translation loss (CTL)—on top of a fixed BGL forecaster for both datasets. The full GlyRAG configuration (RAG + CA + CTL) consistently yields the best or near‑best RMSE/MAE, reducing 60‑min RMSE from 21.59 to 20.22 on Ohio and from 23.52 to 22.32 on AZT1D (5\% improvement over the baseline). Any setting that activates RAG and/or context improves upon the BGL‑only model, but CA+CTL without RAG offers (4th row on ~\tblref{ablation-study}) only modest changes, indicating that cross‑attention over retrieved analogs is necessary to fully exploit the context‑aligned representation. When CTL is used without CA, performance degrades further relative to the full model, suggesting that embedding alignment alone is insufficient in the absence of an explicit context token.

\begin{figure}
\centering
\includegraphics[width=1\linewidth]{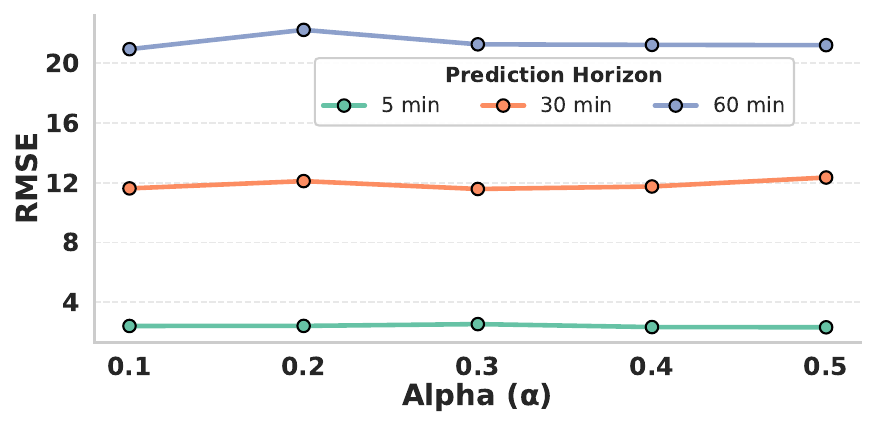}
\vspace{-4mm}
\caption{\textbf{Sensitivity of GlyRAG to $\alpha$.} Mean RMSE across 4 patients at 5-, 30-, and 60-minute horizons as the weighting parameter $\alpha$ varies from 0.1 to 0.5. Performance remains largely insensitive to $\alpha$, with only modest, horizon-dependent fluctuations—indicating robustness of GlyRAG to this hyperparameter.}
\label{fig:ablation_alpha_effect}
\vspace{-5mm}
\end{figure}

~\figref{ablation_alpha_effect} shows the sensitivity of GlyRAG to the translation‑loss weight $\alpha$, plotting mean RMSE across four patients at 5, 30 and 60-min horizons as $\alpha$ varies from 0.1 to 0.5. RMSE curves remain largely flat, with only small horizon‑dependent variations (the 30- and 60-min horizons achieving slightly lower error around $\alpha\approx$0.3), indicating that GlyRAG is not overly sensitive to the exact choice of $\alpha$. This suggests that the cross‑modal translation acts as a stable regularizer: moderate weighting is enough to benefit from the alignment objective without destabilizing forecasting performance.

\section{Discussion and Future Work}
\textcolor{black}{
This study shows that CGM-derived context can improve glucose forecasting when used as an auxiliary representation rather than as a direct clinical decision module. GlyRAG combines morphology-aware LLM summaries with retrieval of similar same-participant training episodes, allowing the forecaster to use both recent CGM dynamics and case-based historical analogues. The strongest gains appear at 30- and 60-minute horizons, where recent CGM continuity alone is less sufficient and contextual information can help capture trend changes, plateaus, and rebounds.}

\textcolor{black}{
The ablation results indicate that the improvement is not explained simply by adding another branch to the model. Handcrafted context variants are inconsistent, whereas the CGM-only LLM narrative provides more stable gains. Retrieval further acts as a secondary, model-dependent refinement: it helps most clearly in the GPT-4 setting but is not uniformly beneficial across all LLMs and horizons. This supports using the CGM-only $M0$ configuration as the main setting. We also examined whether increasing model capacity alone could close the performance gap. Increasing the CNN-RNN baseline capacity by adding additional layers produced less than 5\% RMSE improvement across prediction horizons and did not close the gap to modern time-series baselines or GlyRAG.
}

Our ablation study in Table~\ref{tab:ablation-study} further suggests that context and retrieval are complementary: the best performance arises when cross-attention, cross-translation loss (CTL), and RAG are used together, whereas removing RAG or CTL erodes the gains, and performance remains largely stable across a range of $\alpha$ values—indicating that the translation term behaves as a robust regularizer rather than a fragile tuning knob.

Our work has several limitations that we plan to address in future studies. First, the current evaluation is retrospective and focused on T1D cohorts; broader validation on type 2 diabetes and pre-diabetes populations is needed to assess the generalizability of GlyRAG. Second, context signals can be noisy and incomplete, and the language module is not fine-tuned, which may limit domain faithfulness. The LLM summaries are therefore not clinically validated explanations and should be interpreted as auxiliary model inputs rather than clinician-facing recommendations. \textcolor{black}{ Also, GPT-4 benefits from large-scale pretraining unavailable to the forecasting baselines, which may contribute to its performance advantage; future work should compare against baselines with comparable pretrained representations to isolate this effect. Although robustness across GPT-4 and LLaMA-3.1 was evaluated, sensitivity to alternative valid summaries from the same LLM was not explicitly assessed.} Third, the interpretability provided by retrieved episodes and attention weights is qualitative; future work should quantify explanation fidelity and clinician agreement. Finally, deployment will require prospective studies of real-time robustness, calibration, false-alarm burden, clinical utility, efficient context generation through caching or distilled contextualizers, privacy-preserving retrieval, and uncertainty-aware safeguards. \textcolor{black}{Highly overlapping training windows may produce near-duplicate retrieval candidates; although retrieval is restricted to the training set, we did not quantify sensitivity to temporal exclusion between retrieved windows.}

Future work will: (i) conduct prospective trials with active interventions (e.g., carb advice, temporary basal adjustments, exercise prompts) and with additional sensors/modalities; (ii) scale to larger, multi-site cohorts and diverse diabetes types, including leave-one-subject-out evaluation, population-level pretraining with subject-specific fine-tuning and lightweight calibration through RAG for new subjects; (iii) fine-tune and safety-align the LLM for diabetes discourse; (iv) enrich retrieval with privacy-preserving, context-paired libraries; and (v) strengthen uncertainty quantification to support risk-aware recommendations. These steps move from accurate predictions toward actionable, safe decision support in everyday diabetes care.

\section{Conclusions}
\textcolor{black}
{This paper introduced GlyRAG, a context-aware retrieval-augmented framework for blood glucose forecasting. GlyRAG uses an LLM as a contextualization module to convert CGM windows into morphology-aware summaries, then combines these summaries with numerical glucose embeddings and same-participant retrieval from training episodes.  Across two T1D cohorts, GlyRAG improved long-horizon RMSE over strong CGM-only baselines, with statistically significant gains at 30- and 60-minute horizons, and maintained favorable clinical safety profiles, including a high proportion (85\%) of predictions in clinically acceptable error-grid regions.}

\textcolor{black}{These results provide a foundation for context-aware glucose forecasting systems that can summarize CGM morphology, retrieve relevant patient-specific historical patterns, and support more transparent long-horizon prediction. With prospective validation, uncertainty estimation, and clinical safeguards, such systems may contribute to proactive and human-centered diabetes decision support.}


\section*{References}\vspace{-5mm}
\bibliographystyle{IEEEtran}
\bibliography{ref.bib}

\begin{thebibliography}{10}
\providecommand{\url}[1]{#1}
\csname url@samestyle\endcsname
\providecommand{\newblock}{\relax}
\providecommand{\bibinfo}[2]{#2}
\providecommand{\BIBentrySTDinterwordspacing}{\spaceskip=0pt\relax}
\providecommand{\BIBentryALTinterwordstretchfactor}{4}
\providecommand{\BIBentryALTinterwordspacing}{\spaceskip=\fontdimen2\font plus
\BIBentryALTinterwordstretchfactor\fontdimen3\font minus \fontdimen4\font\relax}
\providecommand{\BIBforeignlanguage}[2]{{%
\expandafter\ifx\csname l@#1\endcsname\relax
\typeout{** WARNING: IEEEtran.bst: No hyphenation pattern has been}%
\typeout{** loaded for the language `#1'. Using the pattern for}%
\typeout{** the default language instead.}%
\else
\language=\csname l@#1\endcsname
\fi
#2}}
\providecommand{\BIBdecl}{\relax}
\BIBdecl

\bibitem{WHOdiabetes2020}
{World Health Organization}, ``Diabetes,'' \url{https://www.who.int/news-room/fact-sheets/detail/diabetes}, 2020, accessed: 2025-01-25.

\bibitem{diabetes2005intensive}
D.~Control, C.~T. of~Diabetes~Interventions, and C.~D. S.~R. Group, ``Intensive diabetes treatment and cardiovascular disease in patients with type 1 diabetes,'' \emph{New England Journal of Medicine}, vol. 353, no.~25, pp. 2643--2653, 2005.

\bibitem{karvonen2000incidence}
M.~Karvonen, M.~Viik-Kajander, E.~Moltchanova, I.~Libman, R.~LaPorte, and J.~Tuomilehto, ``Incidence of childhood type 1 diabetes worldwide. diabetes mondiale (diamond) project group.'' \emph{Diabetes care}, vol.~23, no.~10, pp. 1516--1526, 2000.

\bibitem{idf2021}
{International Diabetes Federation}, ``Idf diabetes atlas,'' \url{https://www.diabetesatlas.org}, 2021, accessed: 2025-01-25.

\bibitem{fullerton2014intensive}
B.~Fullerton, K.~Jeitler, M.~Seitz, K.~Horvath, A.~Berghold, and A.~Siebenhofer, ``Intensive glucose control versus conventional glucose control for type 1 diabetes mellitus,'' \emph{Cochrane Database of Systematic Reviews}, no.~2, 2014.

\bibitem{mccoy2017trajectories}
R.~G. McCoy, C.~Ngufor, H.~K. Van~Houten, B.~Caffo, and N.~D. Shah, ``Trajectories of glycemic change in a national cohort of adults with previously controlled type 2 diabetes,'' \emph{Medical care}, vol.~55, no.~11, pp. 956--964, 2017.

\bibitem{marigliano2024glucose}
M.~Marigliano, C.~Piona, V.~Mancioppi, E.~Morotti, A.~Morandi, and C.~Maffeis, ``Glucose sensor with predictive alarm for hypoglycaemia: Improved glycaemic control in adolescents with type 1 diabetes,'' \emph{Diabetes, Obesity and Metabolism}, vol.~26, no.~4, pp. 1314--1320, 2024.

\bibitem{woldaregay2019data}
A.~Z. Woldaregay, E.~{\AA}rsand, S.~Walderhaug, D.~Albers, L.~Mamykina, T.~Botsis, and G.~Hartvigsen, ``Data-driven modeling and prediction of blood glucose dynamics: Machine learning applications in type 1 diabetes,'' \emph{Artificial intelligence in medicine}, vol.~98, pp. 109--134, 2019.

\bibitem{kwon2025advances}
S.~Y. Kwon and J.~S. Moon, ``Advances in continuous glucose monitoring: clinical applications,'' \emph{Endocrinology and Metabolism}, vol.~40, no.~2, pp. 161--173, 2025.

\bibitem{cappon2019continuous}
G.~Cappon, M.~Vettoretti, G.~Sparacino, and A.~Facchinetti, ``Continuous glucose monitoring sensors for diabetes management: a review of technologies and applications,'' \emph{Diabetes \& metabolism journal}, vol.~43, no.~4, p. 383, 2019.

\bibitem{arefeen2025glytwin}
\BIBentryALTinterwordspacing
A.~Arefeen, S.~Khamesian, M.~A. Grando, B.~Thompson, and H.~Ghasemzadeh, ``Glytwin: Digital twin for glucose control in type 1 diabetes through optimal behavioral modifications using patient-centric counterfactuals,'' 2025. [Online]. Available: \url{https://arxiv.org/abs/2504.09846}
\BIBentrySTDinterwordspacing

\bibitem{xie2020}
J.~Xie and Q.~Wang, ``Benchmarking machine learning algorithms on blood glucose prediction for type i diabetes in comparison with classical time-series models,'' \emph{IEEE Transactions on Biomedical Engineering}, vol.~67, no.~11, pp. 3101--3124, 2020.

\bibitem{mujahid2021}
O.~Mujahid, I.~Contreras, and J.~Vehi, ``Machine learning techniques for hypoglycemia prediction: Trends and challenges,'' \emph{Sensors}, vol.~21, no.~2, p. 546, 2021.

\bibitem{hwang2025generalized}
M.~Hwang, V.~P. Rachim, J.~Yoo, Y.~Lee, and S.-M. Park, ``Generalized multi task learning framework for glucose forecasting and hypoglycemia detection using simulation to reality,'' \emph{npj Digital Medicine}, vol.~8, no.~1, p. 612, 2025.

\bibitem{machiraju2025timeaware}
\BIBentryALTinterwordspacing
A.~Machiraju, E.~Farahmand, S.~B. Soumma, A.~Arefeen, C.~Johnston, and H.~Ghasemzadeh, ``Time-aware cross-attention for multi-modal sensor-based blood glucose forecasting,'' in \emph{IEEE-EMBS International Conference on Body Sensor Networks 2025}, 2025. [Online]. Available: \url{https://openreview.net/forum?id=BYmtjRxfAg}
\BIBentrySTDinterwordspacing

\bibitem{li2019glunet}
K.~Li, C.~Liu, T.~Zhu, P.~Herrero, and P.~Georgiou, ``Glunet: A deep learning framework for accurate glucose forecasting,'' \emph{IEEE journal of biomedical and health informatics}, vol.~24, no.~2, pp. 414--423, 2019.

\bibitem{shuvo2023deep}
M.~M.~H. Shuvo and S.~K. Islam, ``Deep multitask learning by stacked long short-term memory for predicting personalized blood glucose concentration,'' \emph{IEEE Journal of Biomedical and Health Informatics}, vol.~27, no.~3, pp. 1612--1623, 2023.

\bibitem{das2024decoder}
A.~Das, W.~Kong, R.~Sen, and Y.~Zhou, ``A decoder-only foundation model for time-series forecasting,'' in \emph{Forty-first International Conference on Machine Learning}, 2024.

\bibitem{jin2024timellm}
\BIBentryALTinterwordspacing
M.~Jin, S.~Wang, L.~Ma, Z.~Chu, J.~Y. Zhang, X.~Shi, P.-Y. Chen, Y.~Liang, Y.-F. Li, S.~Pan, and Q.~Wen, ``Time-{LLM}: Time series forecasting by reprogramming large language models,'' in \emph{The Twelfth International Conference on Learning Representations}, 2024. [Online]. Available: \url{https://openreview.net/forum?id=Unb5CVPtae}
\BIBentrySTDinterwordspacing

\bibitem{freiburghaus2020deep}
J.~Freiburghaus, A.~Rizzotti, and F.~Albertetti, ``A deep learning approach for blood glucose prediction of type 1 diabetes,'' in \emph{Proceedings of the Proceedings of the 5th International Workshop on Knowledge Discovery in Healthcare Data co-located with 24th European Conference on Artificial Intelligence (ECAI 2020), 29-30 August 2020, Santiago de Compostela, Spain}, vol. 2675.\hskip 1em plus 0.5em minus 0.4em\relax 29-30 August 2020, 2020.

\bibitem{jaloli2023long}
M.~Jaloli and M.~Cescon, ``Long-term prediction of blood glucose levels in type 1 diabetes using a cnn-lstm-based deep neural network,'' \emph{Journal of diabetes science and technology}, vol.~17, no.~6, pp. 1590--1601, 2023.

\bibitem{de2021integration}
M.~De~Bois, M.~A. El-Yacoubi, and M.~Ammi, ``Integration of clinical criteria into the training of deep models: Application to glucose prediction for diabetic people,'' \emph{Smart Health}, vol.~21, p. 100193, 2021.

\bibitem{prendin2023importance}
F.~Prendin, J.~Pavan, G.~Cappon, S.~Del~Favero, G.~Sparacino, and A.~Facchinetti, ``The importance of interpreting machine learning models for blood glucose prediction in diabetes: an analysis using shap,'' \emph{Scientific reports}, vol.~13, no.~1, p. 16865, 2023.

\bibitem{daniels2021multitask}
J.~Daniels, P.~Herrero, and P.~Georgiou, ``A multitask learning approach to personalized blood glucose prediction,'' \emph{IEEE Journal of Biomedical and Health Informatics}, vol.~26, no.~1, pp. 436--445, 2021.

\bibitem{farahmand2025hybridattentionmodelusing}
\BIBentryALTinterwordspacing
E.~Farahmand, S.~B. Soumma, N.~T. Chatrudi, and H.~Ghasemzadeh, ``Hybrid attention model using feature decomposition and knowledge distillation for glucose forecasting,'' 2025. [Online]. Available: \url{https://arxiv.org/abs/2411.10703}
\BIBentrySTDinterwordspacing

\bibitem{garza2023timegpt1}
A.~Garza and M.~Mergenthaler-Canseco, ``Timegpt-1,'' 2023.

\bibitem{xue2023promptcast}
H.~Xue and F.~D. Salim, ``Promptcast: A new prompt-based learning paradigm for time series forecasting,'' \emph{IEEE Transactions on Knowledge and Data Engineering}, vol.~36, no.~11, pp. 6851--6864, 2023.

\bibitem{liu2023large}
X.~Liu, D.~McDuff, G.~Kovacs, I.~Galatzer-Levy, J.~Sunshine, J.~Zhan, M.-Z. Poh, S.~Liao, P.~Di~Achille, and S.~Patel, ``Large language models are few-shot health learners,'' \emph{arXiv preprint arXiv:2305.15525}, 2023.

\bibitem{lee2025timecap}
G.~Lee, W.~Yu, K.~Shin, W.~Cheng, and H.~Chen, ``Timecap: Learning to contextualize, augment, and predict time series events with large language model agents,'' in \emph{Proceedings of the AAAI Conference on Artificial Intelligence}, vol.~39, no.~17, 2025, pp. 18\,082--18\,090.

\bibitem{gokcesu2021generalizedhuberlossrobust}
\BIBentryALTinterwordspacing
K.~Gokcesu and H.~Gokcesu, ``Generalized huber loss for robust learning and its efficient minimization for a robust statistics,'' 2021. [Online]. Available: \url{https://arxiv.org/abs/2108.12627}
\BIBentrySTDinterwordspacing

\bibitem{Devlin2019BERTPO}
\BIBentryALTinterwordspacing
J.~Devlin, M.-W. Chang, K.~Lee, and K.~Toutanova, ``Bert: Pre-training of deep bidirectional transformers for language understanding,'' in \emph{North American Chapter of the Association for Computational Linguistics}, 2019. [Online]. Available: \url{https://api.semanticscholar.org/CorpusID:52967399}
\BIBentrySTDinterwordspacing

\bibitem{PatchTST}
Y.~Nie, N.~H.~Nguyen, P.~Sinthong, and J.~Kalagnanam, ``A time series is worth 64 words: Long-term forecasting with transformers,'' in \emph{International Conference on Learning Representations}, 2023.

\bibitem{marling2020ohiot1dm}
C.~Marling and R.~Bunescu, ``The ohiot1dm dataset for blood glucose level prediction: Update 2020,'' in \emph{CEUR workshop proceedings}, vol. 2675, 2020, p.~71.

\bibitem{AZT1D}
\BIBentryALTinterwordspacing
S.~Khamesian, A.~Arefeen, B.~M. Thompson, A.~Grando, and H.~Ghasemzadeh, ``Azt1d: A real-world dataset for type 1 diabetes,'' 2025. [Online]. Available: \url{https://data.mendeley.com/datasets/gk9m674wcx/1}
\BIBentrySTDinterwordspacing

\bibitem{arefeen2023glysim}
A.~Arefeen and H.~Ghasemzadeh, ``Glysim: Modeling and simulating glycemic response for behavioral lifestyle interventions,'' in \emph{2023 IEEE EMBS International Conference on Biomedical and Health Informatics (BHI)}.\hskip 1em plus 0.5em minus 0.4em\relax IEEE, 2023, pp. 1--5.

\bibitem{glimmer}
\BIBentryALTinterwordspacing
S.~Khamesian, A.~Arefeen, M.~A. Grando, B.~M. Thompson, and H.~Ghasemzadeh, ``Type 1 diabetes management using glimmer: Glucose level indicator model with modified error rate,'' 2025. [Online]. Available: \url{https://arxiv.org/abs/2502.14183}
\BIBentrySTDinterwordspacing

\bibitem{clarkGrid}
\BIBentryALTinterwordspacing
W.~L. Clarke, D.~Cox, L.~A. Gonder-Frederick, W.~Carter, and S.~L. Pohl, ``Evaluating clinical accuracy of systems for self-monitoring of blood glucose,'' \emph{Diabetes Care}, vol.~10, no.~5, pp. 622--628, 09 1987. [Online]. Available: \url{https://doi.org/10.2337/diacare.10.5.622}
\BIBentrySTDinterwordspacing

\end{thebibliography}

\vspace{-14mm}
\begin{IEEEbiography}[{\includegraphics[width=1in,height=1.25in,trim={10 00 40 40}, clip,keepaspectratio]{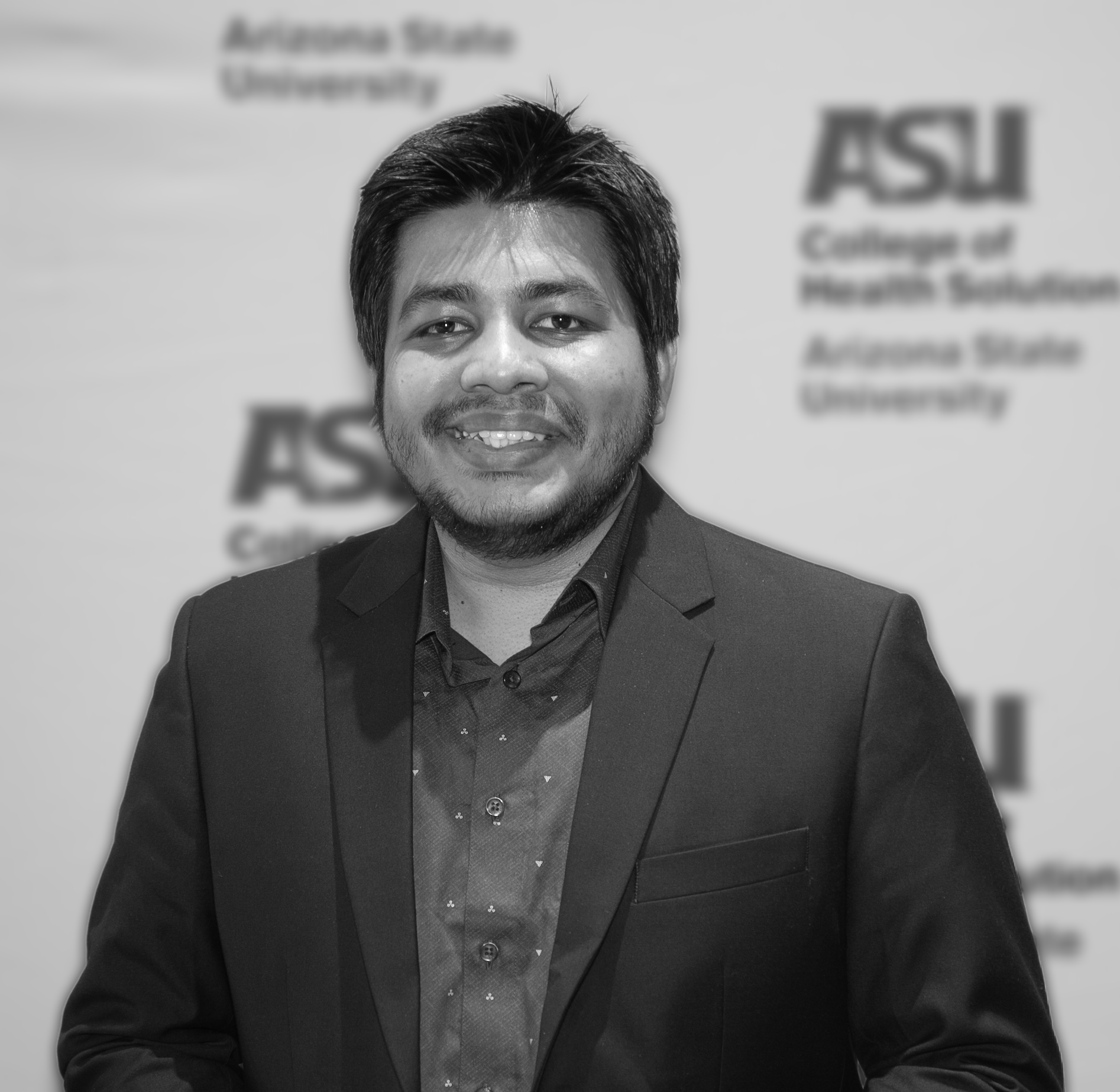}}]{Shovito Barua Soumma, MS} (Student Member, IEEE) received his BSc degree in Computer Science \& Engineering from the Bangladesh University of Engineering \& Technology (BUET), Dhaka, Bangladesh, in 2022 and his MSc degree from Arizona State University in 2025. Currently, he is working toward a PhD degree in Biomedical Informatics \& Data Science at Arizona State University. He is interested in digital health, embedded systems, agentic AI and explainable AI. The focus of his research is on the design and development of power-efficient AI models for wearable monitoring systems with various applications in healthcare. 
\end{IEEEbiography}
\vspace{-12mm}

\begin{IEEEbiography}
[{\includegraphics[width=1in,height=1.25in,clip,keepaspectratio]{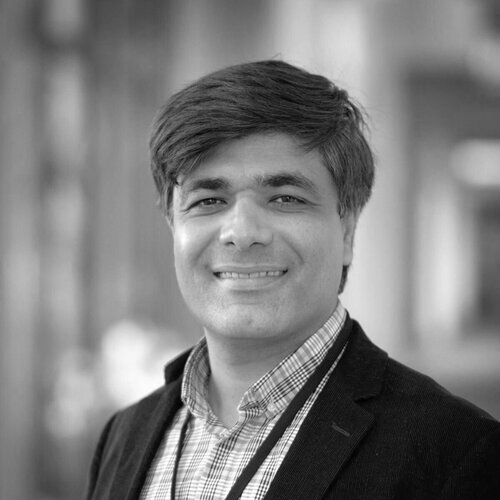}}] {Hassan Ghasemzadeh, PhD} (Senior Member, IEEE), received the BSc degree from the Sharif University of Technology, Tehran, Iran, in 1998, the MSc degree from the University of Tehran, Tehran, Iran, in 2001, and the PhD degree from the University of Texas at Dallas, Richardson, TX, in 2010, all in computer engineering. He was on the faculty of Azad University from 2003-2006 where he served as founding chair of the Computer Science and Engineering Department at the Damavand branch, Tehran, Iran. He spent the academic year 2010- 2011 as a postdoctoral fellow at the West Wireless Health Institute, La Jolla, CA. He was a research manager at the UCLA Wireless Health Institute 2011-2013. Currently he is a Program Director and Professor of Biomedical Informatics \& Data Science, and a graduate faculty of Computer science, Computer engineering, and Biomedical Engineering at Arizona State University (ASU). Prior to joining ASU, he was an Assistant/Associate Professor of Computer Science at Washington State University from 2014 to 2021. The focus of his research is on digital health, machine learning, algorithm design, and system level optimization of embedded and pervasive systems with applications in healthcare and wellness.
\end{IEEEbiography}

\end{document}